\documentclass[]{article}   
\usepackage{graphicx}%
\usepackage{caption}
\usepackage{multirow}%
\usepackage{amsmath,amssymb,amsfonts}%
\usepackage{amsthm}%
\usepackage{mathrsfs}%
\usepackage[title]{appendix}%
\usepackage{xcolor}%
\usepackage{textcomp}%
\usepackage{manyfoot}%
\usepackage{booktabs}%
\usepackage{algorithm}%
\usepackage{algorithmicx}%
\usepackage{algpseudocode}%
\usepackage{listings}%
\usepackage[T1]{fontenc}
\usepackage[utf8]{inputenc}
\usepackage{authblk}
\usepackage{float}
\usepackage{graphicx}
\usepackage{geometry}
\usepackage{cite}
\usepackage{abstract}
\usepackage{lineno}
\usepackage{booktabs}  
\usepackage[table]{xcolor}
\usepackage{multirow}
\usepackage{booktabs}
\usepackage{amssymb}
\usepackage{url}
\usepackage{subcaption}
\usepackage{booktabs}
\usepackage{tabularx}
\usepackage{subcaption}
\usepackage{booktabs}
\usepackage{tabularx}
\usepackage{subcaption}
\usepackage{amssymb}

\usepackage[table]{xcolor}
\definecolor{HeaderColor}{RGB}{215,213,191}
\definecolor{RowA}{RGB}{250,250,246}
\definecolor{RowB}{RGB}{237,235,225}

\title{Towards Personalized Multi-Modal MRI Synthesis across Heterogeneous Datasets}

\author[1,2,3]{Yue Zhang
\thanks{Yue Zhang, Zhizheng Zhuo and Siyao Xu contribute equally to this work.}}
\author[4]{Zhizheng Zhuo$^*$}
\author[4]{Siyao Xu$^*$}
\author[4]{Shan Lv}
\author[4]{Zhaoxi Liu}
\author[4,5]{Jun Qiu}
\author[6]{Qiuli Wang}
\author[4,7]{Yaou Liu$^{\dag}$}

\author[1,2,8,9]{S. Kevin Zhou \thanks{Corresponding authors: Yaou Liu (Yaouliu80@163.com); S. Kevin Zhou (skevinzhou@ustc.edu.cn)}}

\affil[1]{School of Biomedical Engineering, Division of Life Sciences and Medicine, University of Science and Technology of China, Hefei, Anhui, 230026, China}
\affil[2]{Center for Medical Imaging, Robotics, Analytic Computing \& Learning (MIRACLE), Suzhou Institute for Advanced Research, University of Science and Technology of China, Suzhou, Jiangsu, 215123, China}
\affil[3]{National Key Laboratory of Intelligent Collaborative Computing, University of Electronic Science and Technology of China, Chengdu, 610097, China}
\affil[4]{Department of Radiology, Beijing Tiantan Hospital, Capital Medical University, Beijing, 100070, China}
\affil[5]{Department of Radiology, The First Affiliated Hospital of USTC, Division of Life Sciences and Medicine, University of Science and Technology of China, Hefei, Anhui, 230036, China}
\affil[6]{Department of Radiology, Southwest Hospital, Army Medical University, Chongqing, 400038, China}
\affil[7]{China National Clinical Research Center for Neurological Diseases, Beijing, 100070, China}
\affil[8]{State Key Laboratory of Precision and Intelligent Chemistry, University of Science and Technology of China, Hefei, Anhui, 230026, China}
\affil[9]{Jiangsu Provincial Key Laboratory of Multimodal Digital Twin Technology, USTC, Suzhou Jiangsu, 215123, China}

\date{}      

\begin{document}

\captionsetup[figure]{labelfont={bf},labelformat={default},labelsep=period,name={Fig.}}

\maketitle 
\setlength{\absleftindent}{0pt}
\setlength{\absrightindent}{0pt}
\renewcommand{\abstractnamefont}{\fzlishu\bfseries\large}
\renewcommand{\abstracttextfont}{\fzkaiti}

\renewcommand{\topfraction}{0.9}
\renewcommand{\bottomfraction}{0.8}
\renewcommand{\textfraction}{0.05}
\renewcommand{\floatpagefraction}{0.8}


\section*{Abstract}
Synthesizing missing modalities in multi-modal magnetic resonance imaging (MRI) is vital for ensuring diagnostic completeness, particularly when full acquisitions are infeasible due to time constraints, motion artifacts, and patient tolerance. 
Recent unified synthesis models have enabled flexible synthesis tasks by accommodating various input–output configurations. However, their training and evaluation are typically restricted to a single dataset, limiting their generalizability across diverse clinical datasets and impeding practical deployment.
To address this limitation, we propose PMM-Synth, a personalized MRI synthesis framework that not only supports various synthesis tasks but also generalizes effectively across heterogeneous datasets. PMM-Synth is jointly trained on multiple multi-modal MRI datasets that differ in modality coverage, disease types, and intensity distributions. 
It achieves cross-dataset generalization through three core innovations:
a Personalized Feature Modulation module that dynamically adapts feature representations based on dataset identifier to mitigate the impact of distributional shifts; a Modality-Consistent Batch Scheduler that facilitates stable and efficient batch training under inconsistent modality conditions; and a selective supervision loss to ensure effective learning when ground truth modalities are partially missing. Evaluated on four clinical multi-modal MRI datasets, PMM-Synth consistently outperforms state-of-the-art methods in both one-to-one and many-to-one synthesis tasks, achieving superior PSNR and SSIM scores. Qualitative results further demonstrate improved preservation of anatomical structures and pathological details. Additionally, downstream tumor segmentation and radiological reporting studies suggest that PMM-Synth holds potential for supporting reliable diagnosis under real-world modality-missing scenarios.

\section*{Introduction}     
\begin{figure*}[htbp!]
\centering
\includegraphics[width=\textwidth]{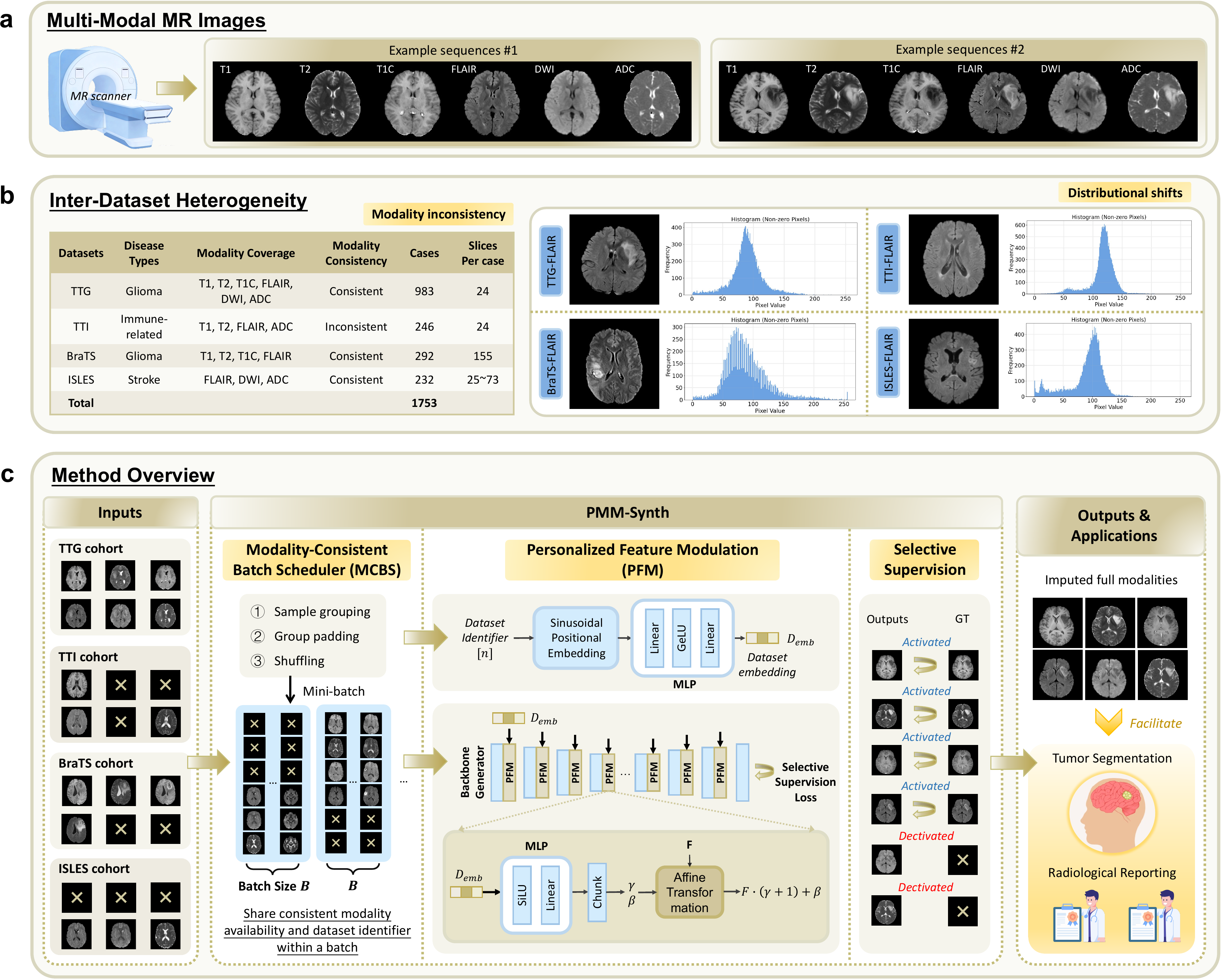}
\caption{Overview of PMM-Synth. (a) Commonly used multi-modal brain MR imaging includes T1, T2, T1C, FLAIR, DWI, and ADC sequences. (b) Illustration of inter-dataset heterogeneity, which mainly includes modality inconsistency and distributional shifts. Modality inconsistency refers to the fact that different datasets cover different modality combinations, and cases within the same dataset may also contain varying subsets of modalities (e.g., the TTI dataset). Distributional shifts are exemplified using the FLAIR modality, where images from different datasets show clear differences in visual appearance and intensity distribution.
(c) Architectural design of PMM-Synth. The framework includes three core components: the PFM to model dataset-specific distributions, the MCBS to enable efficient batch training under modality inconsistency, and the selective supervision loss to enable effective learning with partially missing ground truth. PMM-Synth improves the performance in downstream clinical tasks such as tumor segmentation and radiological reporting.}
\label{fig_overview}
\end{figure*}

Magnetic resonance imaging (MRI) plays a central role in the diagnosis and management of neurological diseases, as it provides non-invasive, high-resolution visualization of brain anatomy and pathology. In clinical practice, multiple MRI sequences are often acquired to capture complementary aspects of tissue characteristics (see Fig.~\ref{fig_overview}a). Among them, T1-weighted imaging (T1) provides excellent anatomical detail and is useful for delineating normal brain structures and fat-containing regions. T2-weighted imaging (T2) is sensitive to water content, aiding in the detection of edema, inflammation, and cystic lesions. Contrast-enhanced T1 (T1C) highlights regions with blood–brain barrier disruption, improving lesion delineation and vascular assessment. Fluid-Attenuated Inversion Recovery (FLAIR) suppresses cerebrospinal fluid signals, enhancing the visibility of periventricular, cortical, and subcortical lesions. Diffusion-Weighted Imaging (DWI) and the derived Apparent Diffusion Coefficient (ADC) maps characterize water diffusion, enabling rapid identification of acute ischemic stroke, assessment of tissue cellularity, and distinction between cytotoxic and vasogenic edema. By integrating these modalities, clinicians can obtain a more comprehensive understanding of brain pathologies, supporting accurate diagnosis, treatment planning, and prognosis evaluation.  
Despite their clinical value, it is often challenging to acquire a complete set of modalities for every patient. Time limitations, patient intolerance, motion artifacts, and contraindications to contrast agents frequently result in missing sequences. For example, stroke patients may undergo urgent DWI but lack contrast-enhanced scans due to time constraints, while tumor patients may have T1, T2, and FLAIR available but lack other modalities.
This incompleteness presents a major challenge for downstream tasks such as tumor segmentation and imaging diagnosis, where the absence of key modalities can impair model performance and reduce diagnostic reliability.

To address the problem of missing modalities, generative models~\cite{goodfellow2020generative,creswell2018generative,10.1145/3626235} have been widely adopted to synthesize unavailable MRI sequences from the available ones. In recent years, research has increasingly focused on improving the \textit{task-level versatility} of synthesis models, aiming to support synthesis tasks with a broad range of input-output configurations. This trend has driven the evolution from one-to-one synthesis~\cite{bowles2016pseudo,roy2016patch,roy2013magnetic,huang2016geometry,jog2015mr,ye2013modality,vemulapalli2015unsupervised,van2015cross,sevetlidis2016whole,li2014deep,nie2017medical,nie2018medical,wang2021realistic,dar2019image,yuan2020unified,luo2022adaptive,fei2022classification,ozbey2023unsupervised,xing2024cross}, to many-to-one synthesis~\cite{jog2014random,jog2017random,lee2019collagan,li2019diamondgan,zhou2020hi,peng2021multi,olut2018generative,yang2019bi,hagiwara2019improving,jiang2023cola}, and more recently to unified synthesis frameworks~\cite{chartsias2017multimodal,sharma2019missing,shen2020multi,dalmaz2022resvit,liu2023one,zhang2024unified} that support flexible synthesis tasks within a single trained model.
Representative unified approaches, such as MM-GAN~\cite{sharma2019missing}, ResViT~\cite{dalmaz2022resvit}, and Uni-Synth~\cite{zhang2024unified}, have demonstrated impressive task-level flexibility. However, these methods are typically trained and evaluated on a single dataset (such as the BraTS dataset~\cite{menze2014multimodal,bakas2017advancing,bakas2018identifying}).
From a clinical perspective, task-level versatility alone is insufficient for real-world deployment. In routine clinical practice, MRI data are acquired across multiple institutions and imaging protocols, leading to substantial inter-dataset heterogeneity. A synthesis model trained on a single curated dataset, even if it supports flexible input-output configurations, often generalize poorly to unseen datasets. 
Consequently, hospitals are compelled to retrain or fine-tune models for each dataset, and model performance remains constrained by the limited scale and diversity of individual datasets.
These limitations highlight the need for \textit{dataset-level versatility} in MRI synthesis models: the ability of a single model to leverage information from multiple datasets and generalize across them without dataset-specific retraining. Such capability is critical for clinical deployment, as it reduces the burden of maintaining multiple models and is expected to deliver stronger synthesis performance.

Here, we pose the following question:
\textit{How can we develop a more powerful synthesis model that not only supports diverse synthesis tasks, but also generalizes effectively across multiple heterogeneous datasets?}
Addressing this challenge is essential for developing clinically deployable synthesis solutions that are robust to the heterogeneity of real-world data. A promising direction is to jointly train the synthesis model on multiple heterogeneous datasets. However, multi-dataset training is not trivial due to substantial inter-dataset heterogeneity, which primarily manifests in two aspects.
First, there exist distributional shifts across datasets. Differences in imaging devices, acquisition protocols, and disease types cause the same modality to exhibit varying intensity distributions and visual characteristics (see Fig.~\ref{fig_overview}b).
These shifts lead multi-dataset training to converge toward an averaged representation, which hinders the modeling of dataset-specific features. As a result, performance on individual datasets becomes suboptimal or even degraded.
Second, there exist significant modality inconsistency. Different datasets often contain different sets of imaging modalities, and even within a single dataset, modality coverage may vary case by case (see Fig.~\ref{fig_overview}b). For instance, the publicly available BraTS dataset for glioma includes T1, T2, T1C, and FLAIR, and each case includes the four modalities. Nevertheless, our private TTI dataset for immune-related brain diseases includes T1, T2, FLAIR, and ADC, and each case only contains a subset of these modalities. Such inconsistency hinders batch training, as varying modality availability across samples disrupts unified input formatting, posing challenges to efficient model optimization and convergence.

In this work, we propose \textbf{PMM-Synth}, a \textbf{P}ersonalized \textbf{M}ulti-\textbf{M}odal MRI \textbf{Synth}esis framework jointly trained on multi-datasets with varying modality coverage, intensity distributions, and disease types. PMM-Synth allows a single trained model to not only support various input-output configurations but also generalize effectively across these datasets, thereby reducing the need for dataset-specific retraining and facilitating scalable deployment in real-world radiological workflows.
Specifically, it is built upon our unified synthesis backbone~\cite{zhang2024unified}, and achieves cross-dataset versatility through three key innovations.
First, to handle the inter-dataset distributional shifts, PMM-Synth introduces a novel Personalized Feature Modulation (PFM) module, which explicitly encodes the dataset identifier and modulates feature representations according to the data source. This mechanism facilitates personalized synthesis tailored to each dataset, and enables the generated images to better match the intensity distribution specific to each dataset.
Then, to tackle inconsistent modality availability across and within datasets, PMM-Synth introduces a Modality-Consistent Batch Scheduler (MCBS). This scheduler ensures that all samples in a training batch share the same available modalities and dataset identifier, allowing for efficient and stable batch training. Finally,
PMM-Synth adopts a selective supervision loss that activates only when ground-truth modalities are available, which ensures effective learning when ground truth modalities are partially missing.
We evaluate PMM-Synth on four heterogeneous multi-modal MRI datasets. 
As shown in Fig.~\ref{fig_overview}b, four datasets differ in modality coverage, distributions and disease focus (including glioma, stroke, and immune-related brain diseases). Experimental results show that PMM-Synth consistently outperforms state-of-the-art methods in both PSNR and SSIM across one-to-one and many-to-one synthesis tasks. Qualitative assessments demonstrate improved synthesis of tumor regions and sharper structural boundaries across all datasets. In downstream applications, the synthesized modalities enhance performance in tumor segmentation and radiological reporting, which reflects the potential of PMM-Synth to support clinical decision-making in scenarios with incomplete imaging modalities.

\section*{Results}
\subsection*{Experimental design}
To comprehensively evaluate the performance of PMM-Synth, we conduct a series of experiments on four heterogeneous multi-modal MRI datasets: TTG, TTI, BraTS~\cite{menze2014multimodal,bakas2017advancing,bakas2018identifying}, and ISLES~\cite{de2024robust,hernandez2022isles}. TTG and TTI are private datasets, while BraTS and ISLES are publicly available. These datasets vary in disease types, modality coverage, and intensity distributions, providing a representative benchmark for multi-dataset synthesis.
Trained on all four datasets, PMM-Synth supports arbitrary synthesis tasks among six MRI modalities (T1, T2, T1C, FLAIR, DWI, and ADC), which corresponds to the union of modality coverage across four datasets.

We first compare PMM-Synth with six representative single-dataset and multi-dataset synthesis methods on various synthesis tasks on each dataset. The single-dataset methods, including \textbf{MM-GAN}~\cite{sharma2019missing}, \textbf{MM-Synth}~\cite{chartsias2017multimodal}, and \textbf{ResViT}~\cite{dalmaz2022resvit}, are trained and evaluated on individual datasets. To evaluate the performance of these methods on different datasets, we train one model per dataset, resulting in four models for each method. The multi-dataset methods, including \textbf{pFLSynth}~\cite{dalmaz2024one}, \textbf{FTN}~\cite{zhou2023fedftn}, and \textbf{PMM-Synth w/o PFM} (an ablated version of our model without PFM), are trained using four datasets and enable a single model to operate across them. It is worth noting that, to the best of our knowledge, no existing MRI synthesis methods have been designed for centralized training across multiple heterogeneous datasets. Therefore, we select two federated learning-based methods, pFLSynth and FTN, as the closest alternatives. Although originally developed under federated learning settings, both methods incorporate architectural components that enable dataset-specific personalization. For fair comparison, we adapt them to a centralized setting and train them on the combined four datasets. The `PMM-Synth w/o PFM' variant is trained directly on the mixed multi-datasets (using MCBS to generate modality-consistent mini-batches), without applying any dataset-specific personalization strategies.
All networks are trained for 200 epochs on a single NVIDIA A100 GPU. The initial learning rate is set to $2 \times 10^{-4}$ and linearly decays to zero after 50 epochs. To quantitatively evaluate the synthesis performance of different methods, we adopt PSNR and SSIM as evaluation metrics.

\begin{table}[t]
\centering
\footnotesize
\caption{Quantitative evaluation of one-to-one MRI synthesis across multiple datasets.}
\label{tb_one2one_all}

\begin{subtable}{\linewidth}
\centering
\caption{PSNR results across the TTG, TTI, BraTS, and ISLES datasets.}
\label{tb_one2one_psnr}
\begin{tabularx}{\linewidth}{XXX|XX} 
\hline
\rowcolor[rgb]{0.843,0.839,0.749}  & \multicolumn{2}{c}{\textbf{TTG}} & \multicolumn{2}{c}{\textbf{TTI}} \\ 
\hline
\rowcolor[rgb]{0.843,0.839,0.749} \textbf{Methods} & \textbf{T1$\rightarrow$T1C} & \textbf{T2$\rightarrow$FLAIR} & \textbf{T1$\rightarrow$T2} & \textbf{T2$\rightarrow$FLAIR} \\ 
\hline
\rowcolor[rgb]{0.98,0.98,0.965} MM-GAN & 27.17±2.03 & 27.93±1.82 & 25.43±2.77 & 27.32±2.34 \\
\rowcolor[rgb]{0.929,0.922,0.882} MM-Synth & 27.17±2.10 & 28.07±2.11 & 25.72±1.63 & 28.05±1.58 \\
\rowcolor[rgb]{0.98,0.98,0.965} ResViT & 26.57±2.05 & 28.29±1.82 & 25.03±1.61 & 27.69±1.65 \\
\rowcolor[rgb]{0.929,0.922,0.882} pFLSyn & 26.45±1.87 & 27.54±1.90 & 26.28±1.49 & 28.46±1.29 \\
\rowcolor[rgb]{0.98,0.98,0.965} FTN & 27.55±2.03 & 28.68±2.05 & 27.55±1.94 & 29.62±1.95 \\
\rowcolor[rgb]{0.929,0.922,0.882} PMM-Synth w/o PFM & 27.46±2.14 & 29.37±1.98 & 27.73±2.28 & 29.13±1.97 \\
\rowcolor[rgb]{0.98,0.98,0.965} PMM-Synth & \textbf{27.72±2.10} & \textbf{29.47±1.98} & \textbf{28.07±2.17} & \textbf{29.86±1.99} \\ 
\hline
\rowcolor[rgb]{0.843,0.839,0.749}  & \multicolumn{2}{c}{\textbf{BraTS}} & \multicolumn{2}{c}{\textbf{ISLES}} \\ 
\hline
\rowcolor[rgb]{0.843,0.839,0.749} \textbf{Methods} & \textbf{T1$\rightarrow$T1C} & \textbf{T2$\rightarrow$FLAIR} & \textbf{FLAIR$\rightarrow$DWI} & \textbf{FLAIR$\rightarrow$ADC} \\ 
\hline
\rowcolor[rgb]{0.98,0.98,0.965} MM-GAN & 25.67±1.38 & 25.76±1.41 & 25.10±1.93 & 22.98±1.47 \\
\rowcolor[rgb]{0.929,0.922,0.882} MM-Synth & 26.09±1.64 & 26.18±1.20 & 25.38±1.38 & 23.04±1.50 \\
\rowcolor[rgb]{0.98,0.98,0.965} ResViT & 25.66±1.20 & 26.01±1.18 & 24.83±1.58 & 22.68±1.33 \\
\rowcolor[rgb]{0.929,0.922,0.882} pFLSyn & 25.51±1.29 & 26.00±1.27 & 24.79±1.46 & 22.54±1.30 \\
\rowcolor[rgb]{0.98,0.98,0.965} FTN & 26.60±1.43 & 26.93±1.31 & \textbf{25.68±1.42} & 23.30±1.56 \\
\rowcolor[rgb]{0.929,0.922,0.882} PMM-Synth w/o PFM & 26.34±1.39 & 26.81±1.42 & 25.29±1.63 & 23.37±1.47 \\
\rowcolor[rgb]{0.98,0.98,0.965} PMM-Synth & \textbf{26.85±1.45} & \textbf{27.23±1.41} & 25.54±1.50 & \textbf{23.48±1.45} \\
\hline
\end{tabularx}
\end{subtable}

\vspace{1em}

\begin{subtable}{\linewidth}
\centering
\caption{SSIM results across the TTG, TTI, BraTS, and ISLES datasets.}
\label{tb_one2one_ssim}
\begin{tabularx}{\linewidth}{XXX|XX}
\hline
\rowcolor[rgb]{0.843,0.839,0.749}  & \multicolumn{2}{c}{\textbf{TTG}} & \multicolumn{2}{c}{\textbf{TTI}} \\ 
\hline
\rowcolor[rgb]{0.843,0.839,0.749} \textbf{Methods} & \textbf{T1$\rightarrow$T1C} & \textbf{T2$\rightarrow$FLAIR} & \textbf{T1$\rightarrow$T2} & \textbf{T2$\rightarrow$FLAIR} \\ 
\hline
\rowcolor[rgb]{0.98,0.98,0.965} MM-GAN & 0.918±0.038 & 0.902±0.026 & 0.903±0.068 & 0.879±0.036 \\ 
\rowcolor[rgb]{0.929,0.922,0.882} MM-Synth & 0.920±0.038 & 0.910±0.029 & 0.922±0.023 & 0.896±0.028 \\ 
\rowcolor[rgb]{0.98,0.98,0.965} ResViT & 0.912±0.039 & 0.915±0.027 & 0.910±0.035 & 0.887±0.035 \\ 
\rowcolor[rgb]{0.929,0.922,0.882} pFLSyn & 0.912±0.039 & 0.901±0.029 & 0.928±0.022 & 0.901±0.027 \\ 
\rowcolor[rgb]{0.98,0.98,0.965} FTN & 0.925±0.035 & 0.920±0.027 & 0.939±0.025 & 0.923±0.025 \\ 
\rowcolor[rgb]{0.929,0.922,0.882} PMM-Synth w/o PFM & 0.928±0.036 & 0.934±0.024 & 0.942±0.028 & 0.925±0.030 \\ 
\rowcolor[rgb]{0.98,0.98,0.965} PMM-Synth & \textbf{0.930±0.034} & \textbf{0.934±0.024} & \textbf{0.945±0.027} & \textbf{0.931±0.030} \\ 
\hline
\rowcolor[rgb]{0.843,0.839,0.749}  & \multicolumn{2}{c}{\textbf{BraTS}} & \multicolumn{2}{c}{\textbf{ISLES}} \\ 
\hline
\rowcolor[rgb]{0.843,0.839,0.749} \textbf{Methods} & \textbf{T1$\rightarrow$T1C} & \textbf{T2$\rightarrow$FLAIR} & \textbf{FLAIR$\rightarrow$DWI} & \textbf{FLAIR$\rightarrow$ADC} \\ 
\hline
\rowcolor[rgb]{0.98,0.98,0.965} MM-GAN & 0.852±0.037 & 0.817±0.031 & 0.867±0.037 & 0.839±0.043 \\ 
\rowcolor[rgb]{0.929,0.922,0.882} MM-Synth & 0.869±0.039 & 0.834±0.029 & 0.869±0.030 & 0.839±0.041 \\ 
\rowcolor[rgb]{0.98,0.98,0.965} ResViT & 0.861±0.036 & 0.836±0.028 & 0.866±0.033 & 0.839±0.040 \\ 
\rowcolor[rgb]{0.929,0.922,0.882} pFLSyn & 0.849±0.037 & 0.827±0.030 & 0.860±0.030 & 0.827±0.040 \\ 
\rowcolor[rgb]{0.98,0.98,0.965} FTN & 0.874±0.034 & 0.851±0.028 & 0.876±0.032 & 0.848±0.041 \\ 
\rowcolor[rgb]{0.929,0.922,0.882} PMM-Synth w/o PFM & 0.872±0.035 & 0.855±0.032 & 0.880±0.029 & 0.856±0.041 \\ 
\rowcolor[rgb]{0.98,0.98,0.965} PMM-Synth & \textbf{0.883±0.035} & \textbf{0.863±0.030} & \textbf{0.885±0.029} & \textbf{0.856±0.039} \\
\hline
\end{tabularx}
\end{subtable}

\end{table}

\begin{table}[t]
\centering
\footnotesize
\caption{Quantitative evaluation of many-to-one MRI synthesis across multiple datasets.}
\label{tb_many2one_all}

\begin{subtable}{\linewidth}
\centering
\caption{PSNR results across the TTG, TTI, BraTS, and ISLES datasets.}
\label{tb_many2one_psnr}
\begin{tabularx}{\linewidth}{XXX|XX} 
\hline
\rowcolor[rgb]{0.843,0.835,0.749}  & \multicolumn{2}{c}{\textbf{TTG}} & \multicolumn{2}{c}{\textbf{TTI}} \\ 
\hline
\rowcolor[rgb]{0.843,0.835,0.749} \textbf{Methods} &
\begin{tabular}[c]{@{}>{\cellcolor[rgb]{0.843,0.835,0.749}}c@{}}\textbf{T1+T2}\\\textbf{$\rightarrow$FLAIR}\end{tabular} &
\begin{tabular}[c]{@{}>{\cellcolor[rgb]{0.843,0.835,0.749}}c@{}}\textbf{T1+T2+FLAIR}\\\textbf{$\rightarrow$T1C}\end{tabular} &
\begin{tabular}[c]{@{}>{\cellcolor[rgb]{0.843,0.835,0.749}}c@{}}\textbf{T1+T2}\\\textbf{$\rightarrow$FLAIR}\end{tabular} &
\begin{tabular}[c]{@{}>{\cellcolor[rgb]{0.843,0.835,0.749}}c@{}}\textbf{T1+T2+FLAIR}\\\textbf{$\rightarrow$ADC}\end{tabular} \\ 
\hline
\rowcolor[rgb]{0.98,0.98,0.965} MM-GAN & 29.40±1.45 & 27.51±2.09 & 28.84±2.04 & 23.96±2.36 \\
\rowcolor[rgb]{0.929,0.922,0.882} MM-Synth & 29.40±1.60 & 27.73±2.13 & 29.24±1.98 & 24.85±1.33 \\
\rowcolor[rgb]{0.98,0.98,0.965} ResViT & 29.60±1.53 & 27.15±2.00 & 28.99±1.87 & 24.00±1.25 \\
\rowcolor[rgb]{0.929,0.922,0.882} Ours w/o PFM & 30.77±1.72 & 28.17±2.24 & 31.01±1.77 & 26.29±1.25 \\
\rowcolor[rgb]{0.98,0.98,0.965} Ours & \textbf{30.90±1.66} & \textbf{28.36±2.22} & \textbf{31.45±2.03} & \textbf{26.41±1.42} \\ 
\hline
\rowcolor[rgb]{0.843,0.835,0.749}  & \multicolumn{2}{c}{\textbf{BraTS}} & \multicolumn{2}{c}{\textbf{ISLES}} \\ 
\hline
\rowcolor[rgb]{0.843,0.835,0.749} \textbf{Methods} &
\begin{tabular}[c]{@{}>{\cellcolor[rgb]{0.843,0.835,0.749}}c@{}}\textbf{T1+T2}\\\textbf{$\rightarrow$FLAIR}\end{tabular} &
\begin{tabular}[c]{@{}>{\cellcolor[rgb]{0.843,0.835,0.749}}c@{}}\textbf{T1+T2+FLAIR}\\\textbf{$\rightarrow$T1C}\end{tabular} &
\begin{tabular}[c]{@{}>{\cellcolor[rgb]{0.843,0.835,0.749}}c@{}}\textbf{FLAIR+DWI}\\\textbf{$\rightarrow$ADC}\end{tabular} &
\begin{tabular}[c]{@{}>{\cellcolor[rgb]{0.843,0.835,0.749}}c@{}}\textbf{FLAIR+ADC}\\\textbf{$\rightarrow$DWI}\end{tabular} \\ 
\hline
\rowcolor[rgb]{0.98,0.98,0.965} MM-GAN & 26.50±1.47 & 26.46±1.41 & 26.98±1.46 & 27.69±1.83 \\
\rowcolor[rgb]{0.929,0.922,0.882} MM-Synth & 26.75±1.50 & 26.91±1.50 & 27.32±1.50 & 28.27±1.60 \\
\rowcolor[rgb]{0.98,0.98,0.965} ResViT & 26.35±1.37 & 26.37±1.20 & 27.17±1.30 & 28.22±2.07 \\
\rowcolor[rgb]{0.929,0.922,0.882} PMM-Synth w/o~PFM & 27.25±1.58 & 27.65±1.58 & 28.32±1.54 & 29.44±1.97 \\
\rowcolor[rgb]{0.98,0.98,0.965} PMM-Synth & \textbf{27.71±1.59} & \textbf{28.13±1.64} & \textbf{28.43±1.51} & \textbf{29.60±1.93} \\
\hline
\end{tabularx}
\end{subtable}

\vspace{1em}

\begin{subtable}{\linewidth}
\centering
\caption{SSIM results across the TTG, TTI, BraTS, and ISLES datasets.}
\label{tb_many2one_ssim}
\begin{tabularx}{\linewidth}{XXX|XX} 
\hline
\rowcolor[rgb]{0.843,0.835,0.749}  & \multicolumn{2}{c}{\textbf{TTG}} & \multicolumn{2}{c}{\textbf{TTI}} \\ 
\hline
\rowcolor[rgb]{0.843,0.835,0.749} \textbf{Methods} &
\begin{tabular}[c]{@{}>{\cellcolor[rgb]{0.843,0.835,0.749}}c@{}}\textbf{T1+T2}\\\textbf{$\rightarrow$FLAIR}\end{tabular} &
\begin{tabular}[c]{@{}>{\cellcolor[rgb]{0.843,0.835,0.749}}c@{}}\textbf{T1+T2+FLAIR}\\\textbf{$\rightarrow$T1C}\end{tabular} &
\begin{tabular}[c]{@{}>{\cellcolor[rgb]{0.843,0.835,0.749}}c@{}}\textbf{T1+T2}\\\textbf{$\rightarrow$FLAIR}\end{tabular} &
\begin{tabular}[c]{@{}>{\cellcolor[rgb]{0.843,0.835,0.749}}c@{}}\textbf{T1+T2+FLAIR}\\\textbf{$\rightarrow$ADC}\end{tabular} \\ 
\hline
\rowcolor[rgb]{0.98,0.98,0.965} MM-GAN & 0.925$\pm$0.019 & 0.924$\pm$0.036 & 0.908$\pm$0.030 & 0.875$\pm$0.051 \\ 
\rowcolor[rgb]{0.929,0.922,0.882} MM-Synth & 0.928$\pm$0.022 & 0.928$\pm$0.036 & 0.919$\pm$0.029 & 0.901$\pm$0.029 \\ 
\rowcolor[rgb]{0.98,0.98,0.965} ResViT & 0.932$\pm$0.021 & 0.922$\pm$0.037 & 0.913$\pm$0.031 & 0.879$\pm$0.037 \\ 
\rowcolor[rgb]{0.929,0.922,0.882} PMM-Synth w/o PFM & 0.949$\pm$0.017 & 0.938$\pm$0.032 & 0.944$\pm$0.023 & \textbf{0.925$\pm$0.025} \\ 
\rowcolor[rgb]{0.98,0.98,0.965} PMM-Synth & \textbf{0.950$\pm$0.016} & \textbf{0.941$\pm$0.031} & \textbf{0.949$\pm$0.022} & 0.923$\pm$0.028 \\ 
\hline
\rowcolor[rgb]{0.843,0.835,0.749}  & \multicolumn{2}{c}{\textbf{BraTS}} & \multicolumn{2}{c}{\textbf{ISLES}} \\ 
\hline
\rowcolor[rgb]{0.843,0.835,0.749} \textbf{Methods} &
\begin{tabular}[c]{@{}>{\cellcolor[rgb]{0.843,0.835,0.749}}c@{}}\textbf{T1+T2}\\\textbf{$\rightarrow$FLAIR}\end{tabular} &
\begin{tabular}[c]{@{}>{\cellcolor[rgb]{0.843,0.835,0.749}}c@{}}\textbf{T1+T2+FLAIR}\\\textbf{$\rightarrow$T1C}\end{tabular} &
\begin{tabular}[c]{@{}>{\cellcolor[rgb]{0.843,0.835,0.749}}c@{}}\textbf{FLAIR+DWI}\\\textbf{$\rightarrow$ADC}\end{tabular} &
\begin{tabular}[c]{@{}>{\cellcolor[rgb]{0.843,0.835,0.749}}c@{}}\textbf{FLAIR+ADC}\\\textbf{$\rightarrow$DWI}\end{tabular} \\ 
\hline
\rowcolor[rgb]{0.98,0.98,0.965} MM-GAN & 0.834$\pm$0.029 & 0.869$\pm$0.031 & 0.926$\pm$0.024 & 0.926$\pm$0.021 \\ 
\rowcolor[rgb]{0.929,0.922,0.882} MM-Synth & 0.847$\pm$0.031 & 0.882$\pm$0.033 & 0.933$\pm$0.021 & 0.934$\pm$0.017 \\ 
\rowcolor[rgb]{0.98,0.98,0.965} ResViT & 0.845$\pm$0.027 & 0.881$\pm$0.028 & 0.934$\pm$0.021 & 0.936$\pm$0.018 \\ 
\rowcolor[rgb]{0.929,0.922,0.882} PMM-Synth w/o PFM & 0.864$\pm$0.031 & 0.896$\pm$0.029 & \textbf{0.948$\pm$0.018} & 0.952$\pm$0.014 \\ 
\rowcolor[rgb]{0.98,0.98,0.965} PMM-Synth & \textbf{0.875$\pm$0.029} & \textbf{0.905$\pm$0.029} & 0.947$\pm$0.017 & \textbf{0.953$\pm$0.014} \\
\hline
\end{tabularx}
\end{subtable}

\end{table}

In addition to qualitative and quantitative analyses of synthesis results, we further design clinical downstream experiments to assess the practical utility of PMM-Synth.
Specifically, we perform glioma and stroke lesion segmentation on the BraTS and ISLES datasets, respectively, to evaluate the fidelity of the synthesized images, particularly in lesion regions.
Besides, we conduct a radiological reporting study, in which two experienced radiologists independently write diagnostic reports based on either real or synthesized sequences. The synthesized sequences are generated under three clinically common modality-missing scenarios: (1) only T1, T2, and FLAIR are available; (2) only T1 and T2 are available; and (3) only T1 is available. 
We then examine whether the synthesized images can reliably support diagnostic decisions by analyzing report similarity and diagnostic accuracy.
Finally, we conduct ablation studies to evaluate the individual contributions of PFM and MCBS to performance improvements and training efficiency, highlighting the importance of each component within the framework.

\begin{figure*}[htbp]
\centering

\begin{subfigure}[t]{0.516\textwidth}
    \centering
    \includegraphics[width=\textwidth]{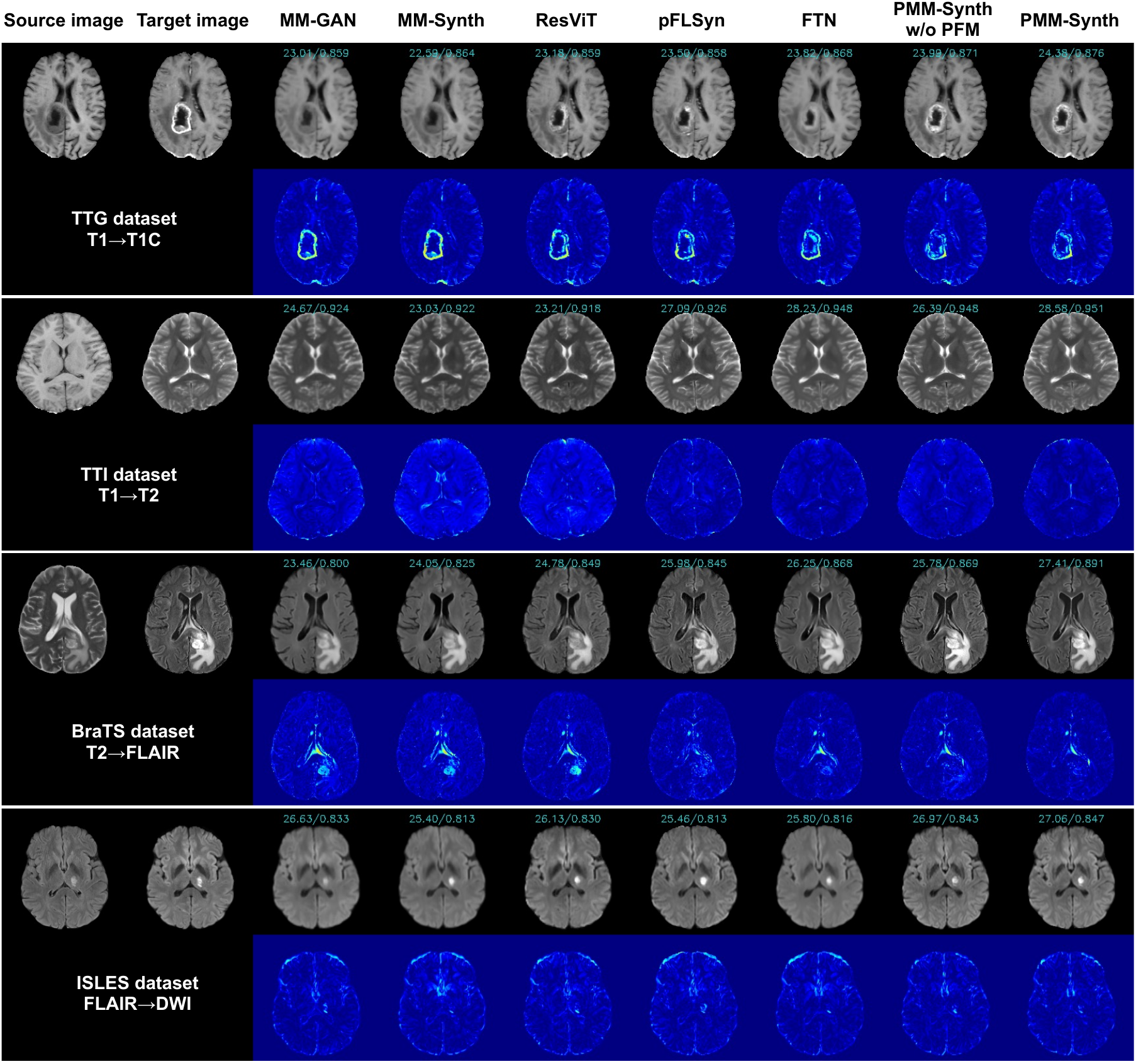}
    \caption{Qualitative comparisons of one-to-one MRI synthesis across heterogeneous datasets. Blue decimals denote PSNR/SSIM values.}
    \label{fig_one2one}
\end{subfigure}
\hfill
\begin{subfigure}[t]{0.464\textwidth}
    \centering
    \includegraphics[width=\textwidth]{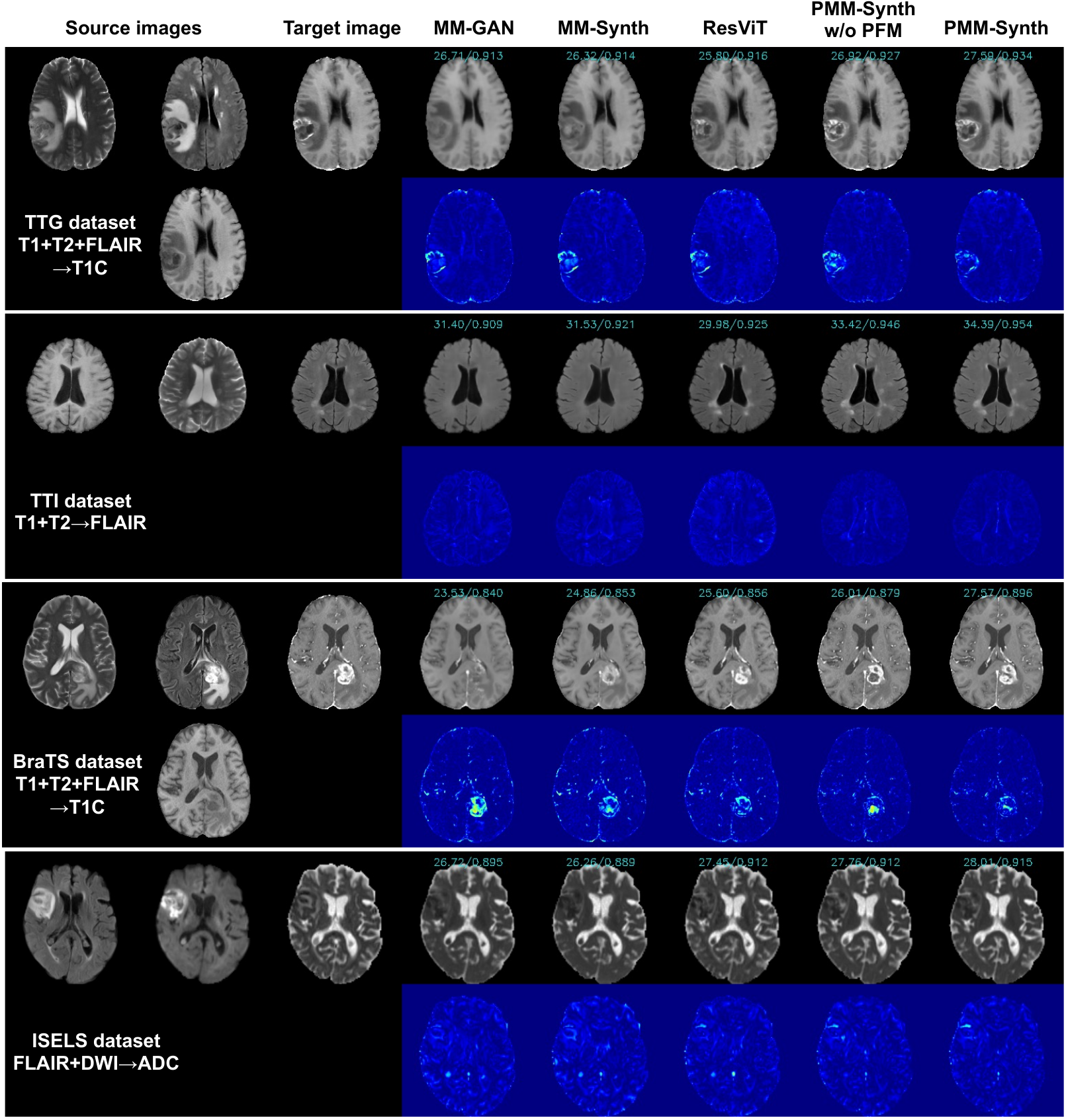}
    \caption{Qualitative comparisons of many-to-one MRI synthesis across heterogeneous datasets. Blue decimals denote PSNR/SSIM values.}
    \label{fig_many2one}
\end{subfigure}

\vspace{0.8em}

\begin{subfigure}[t]{\textwidth}
    \centering
    \includegraphics[width=\textwidth]{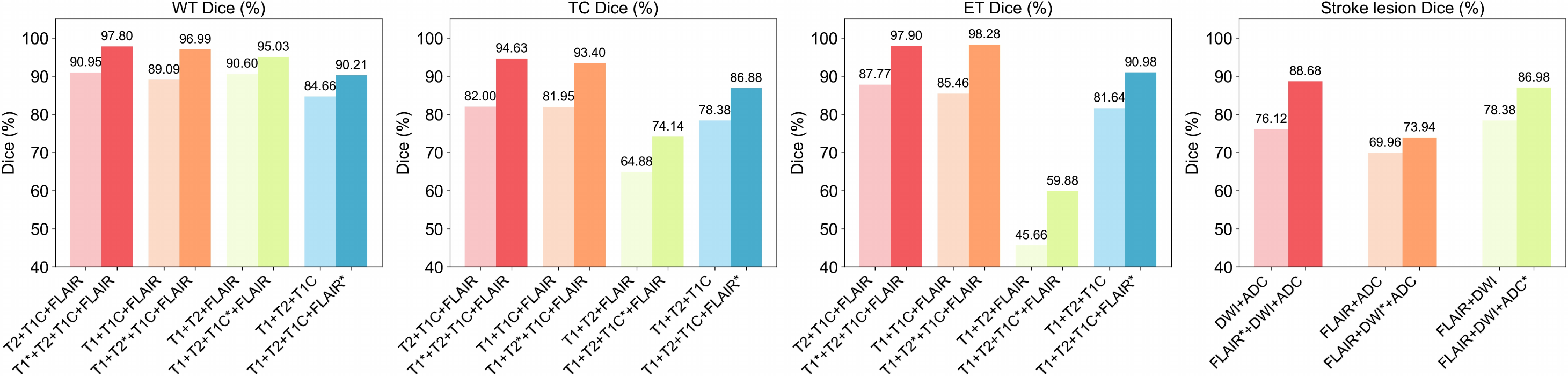}
    \caption{Impact of synthesized modalities on downstream tumor (glioma and stroke lesion) segmentation performance. WT: Whole tumor. TC: Tumor core. ET: Enhancing tumor. Modality marked with * is synthesized by PMM-Synth.}
    \label{fig_seg}
\end{subfigure}

\vspace{0.8em}

\begin{subfigure}[t]{\textwidth}
    \centering
    \includegraphics[width=\textwidth]{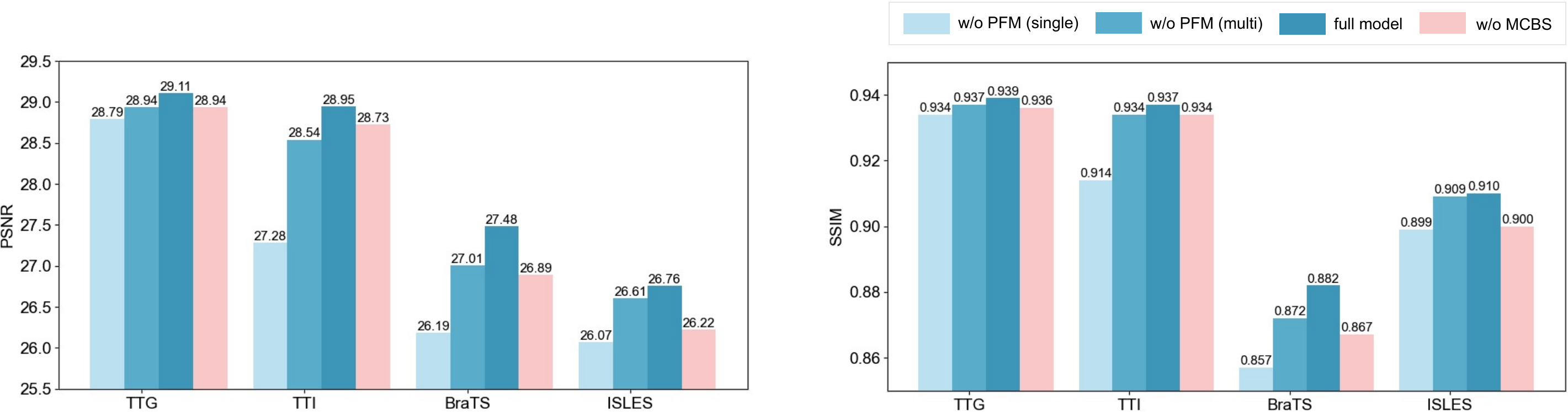}
    \caption{Ablation analysis of PFM module and MCBS module.}
    \label{fig_ablation}
\end{subfigure}

\caption{Comprehensive evaluation and ablation analysis of PMM-Synth across heterogeneous multi-modal MRI datasets.}
\label{fig_synthesis_comparison}
\end{figure*}

\subsection*{PMM-Synth achieves high-fidelity synthesis across heterogeneous datasets}
In this section, we compare PMM-Synth with six single- and multi-dataset synthesis methods to demonstrate its effectiveness. The comparison covers various one-to-one and many-to-one synthesis tasks on four datasets to comprehensively evaluate the performance of each method.

\paragraph{One-to-one synthesis scenarios.}
Since pFLSynth and FTN are one-to-one synthesis models that do not support multi-modal inputs, we first select representative one-to-one synthesis tasks from each dataset for comparison. Specifically, we use T1$\rightarrow$T1C and T2$\rightarrow$FLAIR on the TTG dataset; T1$\rightarrow$T2 and T2$\rightarrow$FLAIR on the TTI dataset; T1$\rightarrow$T1C and T2$\rightarrow$FLAIR on the BraTS dataset; FLAIR$\rightarrow$DWI and FLAIR$\rightarrow$ADC on the ISLES dataset.
Quantitative and qualitative results of one-to-one tasks are presented in Table~\ref{tb_one2one_all} and Fig.~\ref{fig_one2one}. 
Overall, three single-dataset methods exhibit obvious limited performance due to the restricted scale and diversity of a single training dataset.
While FTN trained on multiple datasets achieves competitive quantitative performance, its synthesized images appear noticeably blurred and lack high-frequency details. Besides, FTN is fundamentally limited to one-to-one synthesis tasks and cannot accommodate multi-modal inputs, revealing its restricted task versatility.
Through comparison, PMM-Synth yields the best quantitative results and generates the most visually realistic images across all four datasets.
Especially, PMM-Synth generates clearer tumor enhancement and richer high-frequency details while competing methods often miss or oversmooth. For example, in the row of the TTG dataset in Fig.~\ref{fig_one2one}, MM-GAN, MM-Synth and FTN fail to enhance the tumor region, and pFLSyn produces inaccurate enhancement. In contrast, PMM-Synth preserves realistic texture and edge sharpness, yielding higher fidelity in pathological regions.
Similar advantages are observed in the BraTS dataset, where PMM-Synth generates FLAIR images with sharper lesion contours and more clearer edema areas.
Beyond capturing finer structural details, PMM-Synth also better preserves the intensity characteristics of the target modality, which benefits from its dataset-specific personalization strategy. As shown in the TTI dataset example in Fig.~\ref{fig_one2one}, removing PFM from PMM-Synth (PMM-Synth w/o PFM) leads to more obvious intensity shifts from the real image, particularly in white matter and ventricular regions. 
Such findings confirm that our complete model is robust to distribution variations across datasets and capable of learning dataset-specific characteristics.

\paragraph{Many-to-one synthesis scenarios.}
Except for pFLSyn and FTN, all the other methods are capable of handling synthesis tasks with various input-output configurations. Therefore, we further compare and analyze their performance on many-to-one synthesis tasks.
Specifically, we select T1+T2$\rightarrow$FLAIR and T1+T2+FLAIR$\rightarrow$T1C on the TTG dataset; T1+T2$\rightarrow$FLAIR and T1+T2+FLAIR$\rightarrow$ADC on the TTI dataset; T1+T2$\rightarrow$FLAIR, T1+T2+FLAIR$\rightarrow$T1C on the BraTS dataset; FLAIR+DWI$\rightarrow$ADC, FLAIR+ADC$\rightarrow$DWI on the ISLES dataset.
Table~\ref{tb_many2one_all} and Fig.~\ref{fig_many2one} report the comparison results. PMM-Synth effectively leverages complementary information from multi-modal inputs to enhance image fidelity, and achieves the best quantitative scores and visual quality across all four datasets. For example, the T1+T2+FLAIR$\rightarrow$T1C task on the TTG dataset (in Table~\ref{tb_many2one_all}) outperforms the one-to-one T1$\rightarrow$T1C task (in Table~\ref{tb_one2one_all}) in all quantitative metrics, which is consistent with the expected benefits of multi-modal integration. Results across both one-to-one and many-to-one scenarios demonstrate that PMM-Synth consistently outperforms competing methods on heterogeneous multi-modal MRI datasets, highlighting its superior generalizability and dataset-level versatility.

\subsection*{PMM-Synth facilitates clinical downstream tasks}
Previous sections have demonstrated the superior synthesis quality of PMM-Synth through both quantitative metrics and visual assessments. We now turn to evaluating its impact on downstream clinical tasks. Specifically, we assess whether the synthesized modalities can effectively support two critical applications: tumor segmentation and radiological reporting. These tasks directly reflect the clinical utility of synthesized images in aiding treatment planning through accurate tumor delineation and supporting diagnostic decision-making when certain modalities are unavailable.

\paragraph{Tumor segmentation.}
Based on the disease types included in the datasets, we select glioma segmentation on the BraTS dataset and stroke lesion segmentation on the ISLES dataset. For the BraTS dataset, we consider four scenarios with one modality missing: T2+T1C+FLAIR, T1+T1C+FLAIR, T1+T2+FLAIR, and T1+T2+T1C as available modalities. PMM-Synth is used to synthesize the missing fourth modality (denoted by $*$ in Fig.~\ref{fig_seg}). We train a U-Net accepting four-modal inputs for glioma segmentation, and three U-Nets accepting three-modal inputs to serve as baselines. We report Dice scores for whole tumor (WT), tumor core (TC), and enhancing tumor (ET) by comparing segmentation masks generated from four-modal synthesized sequences (or three-modal real sequences) with four-modal real sequences. 
Similarly, for the ISLES dataset, we consider three scenarios with one modality missing: DWI+ADC, FLAIR+ADC, and FLAIR+DWI as available modalities. PMM-Synth is used to generate the missing third modality. Accordingly, a U-Net that accepts three-modal inputs and two U-Nets that accept two-modal inputs are trained for stroke lesion segmentation.
As illustrated in Fig.~\ref{fig_seg}, PMM-Synth consistently improves segmentation accuracy of WT, TC, and ET compared to three-modal counterparts. This indicates that the synthesized modality effectively recovers missing contrast information beneficial for glioma segmentation.
For stroke lesion segmentation, similar improvements are observed on the ISLES dataset. In particular, the FLAIR+DWI+ADC* setting yields a notable increase in Dice score over its two-modal counterpart (FLAIR+DWI), suggesting that PMM-Synth successfully captures diffusion-relevant features that are critical for stroke characterization.
These findings demonstrate the value of PMM-Synth in completing missing clinical data, which yields tangible benefits for downstream segmentation performance.

\begin{figure*}[t]
\centering
\includegraphics[width=\textwidth]{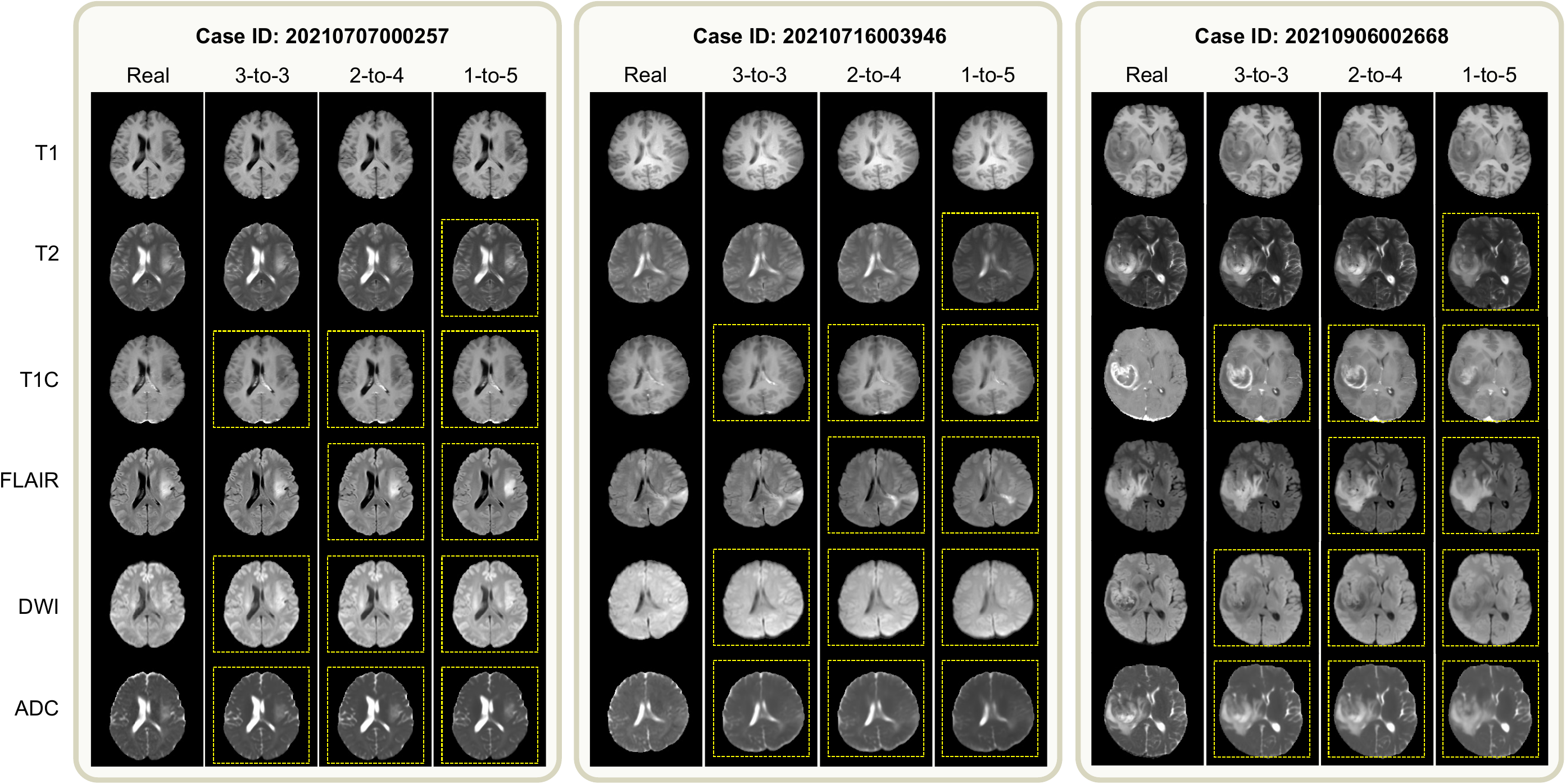}
\caption{Examples of cases with real and synthesized sequences. `3-to-3' refers to synthesizing T1C, DWI, and ADC from T1, T2, and FLAIR; `2-to-4' uses T1 and T2 to generate the remaining four modalities; `1-to-5' uses only T1 to synthesize the other five modalities. The synthesized modalities are highlighted with \textcolor{yellow}{yellow} dashed boxes.}
\label{fig_report_img}
\end{figure*}

\begin{table}[htbp]
  \centering
  \tiny
  \caption{Examples of radiological reports written by two radiologists. Each report consists of the \emph{Findings} and \emph{Impression}. The differences between each report and the ground truth are marked in \textcolor{red}{red}. \textcolor{red}{$\Box\Box\Box$} indicates missing findings/impression compared to the ground truth.}
  \label{tb_reports}

\begin{subtable}{\linewidth}
\centering
\caption{Radiological reports written by Radiologist 1.}
\begin{tabularx}{\linewidth}{XXXXX}
\toprule
\rowcolor[rgb]{0.843,0.839,0.749}
& \textbf{Real sequences}
& \textbf{3-to-3 sequences}
& \textbf{2-to-4 sequences}
& \textbf{1-to-5 sequences} \\
\midrule

\rowcolor[rgb]{0.98,0.98,0.965}
\textbf{Case ID:} 
\newline
20210707000257
&
\textbf{Findings:}\newline
Patchy low signal on T1WI and high signal on T2WI and FLAIR are observed in the left insular lobe, with ill-defined margins. The lesion measures approximately 49 mm × 22 mm on the axial section. No significant diffusion restriction is noted on DWI/ADC. Post-contrast images show small patchy areas of marked enhancement locally.\newline
\textbf{Impression:}\newline
Diffuse glioma, grade III astrocytoma/oligodendroglioma?\newline
&
\textbf{Findings:}\newline
Patchy low signal on T1WI and high signal on T2WI and FLAIR are observed in the left insular lobe, with ill-defined margins. The lesion measures approximately 49 mm × 22 mm on the axial section. No significant diffusion restriction is noted on DWI/ADC. Post-contrast images show small patchy areas of marked enhancement locally. \newline
\textbf{Impression:}\newline
Diffuse glioma, \textcolor{red}{$\square\square\square$}\newline
&
\textbf{Findings:}\newline
Patchy low signal on T1WI and high signal on T2WI and FLAIR are observed in the left insular lobe, with ill-defined margins. The lesion measures approximately 49 mm × 22 mm on the axial section. No significant diffusion restriction is noted on DWI/ADC. Post-contrast images show small patchy areas of marked enhancement locally.\newline
\textbf{Impression:}\newline
Diffuse glioma, \textcolor{red}{$\square\square\square$}\newline
&
\textbf{Findings:}\newline
Patchy low signal on T1WI and high signal on T2WI and FLAIR are observed in the left insular lobe, \textcolor{red}{with relatively well-defined margins}. The lesion measures approximately \textcolor{red}{36 mm × 19 mm} on the axial section. No significant diffusion restriction is noted on DWI/ADC. \textcolor{red}{No definite enhancement is observed on post-contrast images.}\newline
\textbf{Impression:}\newline
\textcolor{red}{Low-grade glioma, $\square\square\square$}\newline
\\[1ex]

\rowcolor[rgb]{0.929,0.922,0.882}
\textbf{Case ID:} 
\newline
20210915004134
&
\textbf{Findings:}\newline
Patchy low signal on T1WI and high signal on T2WI and FLAIR are observed in the left frontal lobe, with relatively well-defined margins. The lesion measures approximately 35 mm × 26 mm on the axial section. No definite diffusion restriction is observed on DWI/ADC, and no obvious enhancement is seen on contrast-enhanced scans.\newline
\textbf{Impression:}\newline
Low-grade glioma, oligodendroglioma?\newline
&
\textbf{Findings:}\newline
Patchy low signal on T1WI and high signal on T2WI and FLAIR are observed in the left frontal lobe, with relatively well-defined margins. The lesion measures approximately 35 mm × 26 mm on the axial section. No definite diffusion restriction is observed on DWI/ADC, and no obvious enhancement is seen on contrast-enhanced scans.\newline
\textbf{Impression:}\newline
\textcolor{red}{$\square\square\square$, }Oligodendroglioma? \textcolor{red}{Astrocytoma?}\newline
&
\textbf{Findings:}\newline
Patchy low signal on T1WI and high signal on T2WI and FLAIR are observed in the left frontal lobe, with relatively well-defined margins. The lesion measures approximately 35 mm × 26 mm on the axial section. No definite diffusion restriction is observed on DWI/ADC, and no obvious enhancement is seen on contrast-enhanced scans.\newline
\textbf{Impression:}\newline
\textcolor{red}{$\square\square\square$, }Oligodendroglioma? \textcolor{red}{Astrocytoma?}\newline
&
\textbf{Findings:}\newline
Patchy low signal on T1WI and high signal on T2WI and FLAIR are observed in the left frontal lobe, with relatively well-defined margins. The lesion measures approximately 35 mm × 26 mm on the axial section. No definite diffusion restriction is observed on DWI/ADC, and no obvious enhancement is seen on contrast-enhanced scans.\newline
\textbf{Impression:}\newline
\textcolor{red}{$\square\square\square$, }Oligodendroglioma? \textcolor{red}{Astrocytoma?}\newline
\\
\bottomrule
\end{tabularx}
\end{subtable}

\vspace{1em}

\begin{subtable}{\linewidth}
\centering
\caption{Radiological reports written by Radiologist 2.}
\begin{tabularx}{\linewidth}{XXXXX}
\toprule
\rowcolor[rgb]{0.843,0.839,0.749}
& \textbf{Real sequences}
& \textbf{3-to-3 sequences}
& \textbf{2-to-4 sequences}
& \textbf{1-to-5 sequences} \\
\midrule

\rowcolor[rgb]{0.98,0.98,0.965}
\textbf{Case ID:} 
\newline
20210707000257
&
\textbf{Findings:}\newline
Patchy low signal on T1WI and high signal on T2WI are present in the left frontal temporal insular region, with high signal on FLAIR. Small round foci with iso signal on T1WI/high signal on T2WI/high signal on FLAIR are observed locally and no diffusion restriction is observed on DWI/ADC. Contrast enhanced imaging shows small patchy areas of relatively homogeneous enhancement, with indistinct margins. The lesion measures approximately 51 mm × 24 mm on the axial section. The left lateral ventricle is compressed and narrowed. The midline remains central.\newline
\textbf{Impression:}\newline
Low-grade glioma\newline
&
\textbf{Findings:}\newline
Patchy low signal on T1WI and high signal on T2WI are present in the left frontal temporal insular region, with high signal on FLAIR. Small round foci with iso signal on T1WI/high signal on T2WI/high signal on FLAIR are observed locally and no diffusion restriction is observed on DWI/ADC. Contrast enhanced imaging shows small patchy areas of \textcolor{red}{heterogeneous enhancement}, with indistinct margins. The lesion measures approximately 51 mm × 24 mm on the axial section. The left lateral ventricle is compressed and narrowed. The midline remains central.\newline
\textbf{Impression:}\newline
Low-grade glioma\newline
&
\textbf{Findings:}\newline
Patchy low signal on T1WI and high signal on T2WI are present in the left frontal temporal insular region, with high signal on FLAIR. Small round foci with iso signal on T1WI/high signal on T2WI/high signal on FLAIR are observed locally and no diffusion restriction is observed on DWI/ADC. Contrast enhanced imaging shows small patchy areas of \textcolor{red}{heterogeneous enhancement}, with indistinct margins. The lesion measures approximately 51 mm × 24 mm on the axial section. The left lateral ventricle is compressed and narrowed. The midline remains central.\newline
\textbf{Impression:}\newline
Low-grade glioma\newline
&
\textbf{Findings:}\newline
Patchy low signal on T1WI and high signal on T2WI are present in the left frontal temporal insular region, with high signal on FLAIR. \textcolor{red}{DWI /ADC demonstrate mildly increased signal.} Contrast enhanced imaging shows small patchy areas of \textcolor{red}{heterogeneous enhancement}, with indistinct margins. The lesion measures approximately \textcolor{red}{40 mm × 23 mm} on the axial section. The left lateral ventricle is compressed and narrowed. The midline remains central.\newline
\textbf{Impression:}\newline
Low-grade glioma\newline
\\[1ex]

\rowcolor[rgb]{0.929,0.922,0.882}
\textbf{Case ID:} 
\newline
20210915004134
&
\textbf{Findings:}\newline
Patchy low signal on T1WI and high signal on T2WI are present in the left frontal lobe, with iso- to high signal on FLAIR. The margins are indistinct. The lesion measures approximately 36 mm × 26 mm. No apparent diffusion restriction is seen on DWI/ADC, and no obvious enhancement is observed on contrast-enhanced imaging.\newline
\textbf{Impression:}\newline
Low-grade glioma, astrocytoma?\newline
&
\textbf{Findings:}\newline
Patchy low signal on T1WI and high signal on T2WI are present in the left frontal lobe, with iso- to high signal on FLAIR. The margins are indistinct. The lesion measures approximately \textcolor{red}{35 mm × 28 mm}. No apparent diffusion restriction is seen on DWI/ADC, and no obvious enhancement is observed on contrast-enhanced imaging.\newline
\textbf{Impression:}\newline
Low-grade glioma, astrocytoma?\newline
&
\textbf{Findings:}\newline
Patchy low signal on T1WI and high signal on T2WI are present in the left frontal lobe, with iso- to high signal on FLAIR. The margins are indistinct. The lesion measures approximately \textcolor{red}{35 mm × 28 mm}. No apparent diffusion restriction is seen on DWI/ADC, and no obvious enhancement is observed on contrast-enhanced imaging.\newline
\textbf{Impression:}\newline
Low-grade glioma, astrocytoma?\newline
&
\textbf{Findings:}\newline
Patchy low signal on T1WI and high signal on T2WI are present in the left frontal lobe, with iso- to high signal on FLAIR. The margins are indistinct. The lesion measures approximately \textcolor{red}{40 mm × 28 mm}. No apparent diffusion restriction is seen on DWI/ADC, and no obvious enhancement is observed on contrast-enhanced imaging.\newline
\textbf{Impression:}\newline
Low-grade glioma, \textcolor{red}{$\square\square\square$}\newline
\\
\bottomrule
\end{tabularx}
\end{subtable}
\end{table}

\begin{table*}[t]
\centering
\footnotesize
\caption{Radiological reporting analysis on the TTG dataset.}
\label{tb_meta_analysis}

\begin{subtable}{\textwidth}
\centering
\caption{Assessment of textual similarity within the \emph{Findings} descriptions.}
\label{tb_meta_analysis_sim}
\begin{tabularx}{\linewidth}{XXXXX}
    \hline
\rowcolor[rgb]{0.843,0.839,0.749}
 & \multicolumn{1}{l}{\textbf{3-to-3 sequences}} & \multicolumn{1}{l}{\textbf{2-to-4 sequences}} & \multicolumn{1}{l}{\textbf{1-to-5 sequences}} \\ 
\hline
\rowcolor[rgb]{0.98,0.98,0.965}
Radiologist 1 & 0.974~ & 0.972~ & 0.843~ \\
\rowcolor[rgb]{0.929,0.922,0.882}
Radiologist 2 & 0.871~ & 0.893~ & 0.860~ \\
\hline
\end{tabularx}
\end{subtable}

\vspace{1em}
\begin{subtable}{\textwidth}
\centering
\caption{Analysis of diagnostic accuracy based on the \emph{Impression} provided by Radiologist 1. LGG: low-grade glioma. HGG: high-grade glioma. ODG: oligodendroglioma. GBM: glioblastoma. EPN: ependymoma. The `unspecified glioma' category includes cases labeled as `glioma' or `diffuse glioma' that lack explicit grading or subtype. The `other' category includes non-neoplastic findings such as `no definite abnormality', `inflammatory lesions', and `imaging artifacts'.}
\label{tb_meta_analysis_acc_1}
\begin{tabularx}{\linewidth}{llXXlXXlXX}
\hline
\rowcolor[rgb]{0.843,0.839,0.749}
\textbf{} & \multicolumn{3}{c}{\textbf{3-to-3 sequences}} & 
\multicolumn{3}{c}{\textbf{2-to-4 sequences}} & 
\multicolumn{3}{c}{\textbf{1-to-5 sequences}} \\
\rowcolor[rgb]{0.843,0.839,0.749}
\textbf{Type} & \textbf{Precision} & \textbf{Recall} & \textbf{F1} &
\textbf{Precision} & \textbf{Recall} & \textbf{F1} &
\textbf{Precision} & \textbf{Recall} & \textbf{F1} \\
\hline

\rowcolor[rgb]{0.98,0.98,0.965}
LGG & 1.000 & 0.750 & 0.857 & 1.000 & 0.750 & 0.857 & 0.667 & 0.500 & 0.571 \\

\rowcolor[rgb]{0.929,0.922,0.882}
HGG & 1.000 & 0.889 & 0.941 & 1.000 & 0.889 & 0.941 & 1.000 & 0.889 & 0.941 \\

\rowcolor[rgb]{0.98,0.98,0.965}
Astrocytoma & 0.875 & 0.875 & 0.875 & 0.875 & 0.875 & 0.875 & 0.875 & 0.875 & 0.875 \\

\rowcolor[rgb]{0.929,0.922,0.882}
ODG & 1.000 & 0.857 & 0.923 & 1.000 & 0.857 & 0.923 & 0.857 & 0.857 & 0.857 \\

\rowcolor[rgb]{0.98,0.98,0.965}
GBM & 1.000 & 1.000 & 1.000 & 0.500 & 1.000 & 0.667 & 1.000 & 1.000 & 1.000 \\

\rowcolor[rgb]{0.929,0.922,0.882}
EPN & 0.500 & 1.000 & 0.667 & 0.500 & 1.000 & 0.667 & 1.000 & 1.000 & 1.000 \\

\rowcolor[rgb]{0.98,0.98,0.965}
Unspecified Glioma & 0.667 & 1.000 & 0.800 & 0.667 & 1.000 & 0.800 & 1.000 & 1.000 & 1.000 \\

\rowcolor[rgb]{0.929,0.922,0.882}
Metastasis & 1.000 & 0.500 & 0.667 & 0.500 & 0.500 & 0.500 & 1.000 & 0.500 & 0.667 \\

\rowcolor[rgb]{0.98,0.98,0.965}
Other & 1.000 & 1.000 & 1.000 & 1.000 & 1.000 & 1.000 & 0.500 & 1.000 & 0.667 \\

\hline

\rowcolor[rgb]{0.929,0.922,0.882}
Macro-average & 0.894 & 0.875 & 0.859 & 0.782 & 0.875 & 0.803 & 0.878 & 0.847 & 0.842 \\

\rowcolor[rgb]{0.98,0.98,0.965}
Micro-average & 0.912 & 0.861 & 0.886 & 0.838 & 0.861 & 0.849 & 0.882 & 0.833 & 0.857 \\
\hline
\end{tabularx}
\end{subtable}

\vspace{1em}

\begin{subtable}{\textwidth}
\centering
\caption{Analysis of diagnostic accuracy based on the \emph{Impression} provided by Radiologist 2.}
\label{tb_meta_analysis_acc_2}
\begin{tabularx}{\linewidth}{llXXlXXlXX}
\hline
\rowcolor[rgb]{0.843,0.839,0.749}
\textbf{} & \multicolumn{3}{c}{\textbf{3-to-3 sequences}} &
\multicolumn{3}{c}{\textbf{2-to-4 sequences}} &
\multicolumn{3}{c}{\textbf{1-to-5 sequences}} \\
\rowcolor[rgb]{0.843,0.839,0.749}
\textbf{Type} & \textbf{Precision} & \textbf{Recall} & \textbf{F1} &
\textbf{Precision} & \textbf{Recall} & \textbf{F1} &
\textbf{Precision} & \textbf{Recall} & \textbf{F1} \\
\hline

\rowcolor[rgb]{0.98,0.98,0.965}
LGG & 1.000 & 0.857 & 0.923 & 1.000 & 1.000 & 1.000 & 1.000 & 1.000 & 1.000 \\

\rowcolor[rgb]{0.929,0.922,0.882}
HGG & 1.000 & 1.000 & 1.000 & 1.000 & 1.000 & 1.000 & 1.000 & 1.000 & 1.000 \\

\rowcolor[rgb]{0.98,0.98,0.965}
Astrocytoma & 1.000 & 1.000 & 1.000 & 1.000 & 1.000 & 1.000 & 1.000 & 0.500 & 0.667 \\

\rowcolor[rgb]{0.929,0.922,0.882}
ODG & / & / & / & / & / & / & / & / & / \\

\rowcolor[rgb]{0.98,0.98,0.965}
GBM & / & / & / & / & / & / & / & / & / \\

\rowcolor[rgb]{0.929,0.922,0.882}
EPN & 1.000 & 1.000 & 1.000 & 1.000 & 1.000 & 1.000 & 1.000 & 1.000 & 1.000 \\

\rowcolor[rgb]{0.98,0.98,0.965}
Unspecified Glioma & 1.000 & 1.000 & 1.000 & 1.000 & 1.000 & 1.000 & 1.000 & 1.000 & 1.000 \\

\rowcolor[rgb]{0.929,0.922,0.882}
Metastasis & 0.500 & 1.000 & 0.667 & 1.000 & 1.000 & 1.000 & 1.000 & 1.000 & 1.000 \\

\rowcolor[rgb]{0.98,0.98,0.965}
Other & 0.750 & 1.000 & 0.857 & 0.750 & 1.000 & 0.857 & 1.000 & 1.000 & 1.000 \\

\hline

\rowcolor[rgb]{0.929,0.922,0.882}
Macro-average & 0.893 & 0.980 & 0.921 & 0.964 & 1.000 & 0.980 & 1.000 & 0.929 & 0.952 \\

\rowcolor[rgb]{0.98,0.98,0.965}
Micro-average & 0.929 & 0.963 & 0.945 & 0.964 & 1.000 & 0.982 & 1.000 & 0.926 & 0.962 \\
\hline
\end{tabularx}
\end{subtable}

\end{table*}

\paragraph{Radiological reporting study.}
While the segmentation results demonstrate PMM-Synth's effectiveness in supporting accurate tumor delineation, its potential for clinical application ultimately depends on whether the synthesized images can meaningfully support clinical interpretation. To this end, we conduct a controlled radiological reporting study. Two board-certified radiologists from Beijing Tiantan Hospital, with six to ten years of clinical experience, are invited to independently review MRI sequences and produce structured diagnostic reports, enabling a direct comparison between reports derived from real acquisitions and synthetic ones.

Our report analysis primarily focus on glioma cases. Specifically, 20 cases are randomly selected from the TTG cohort for detailed evaluation. For each case, four sets of image sequences are constructed to form a within-case controlled comparison. These include one fully real reference set and three synthetic sets generated under different modality-missing scenarios (see Fig.~\ref{fig_report_img}). The real set contain all six modalities (T1, T2, T1C, FLAIR, DWI, and ADC) and serve as the ground-truth. The three synthesis settings progressively reduce the number of real input modalities while completing the missing sequences using PMM-Synth: (1) a \textbf{3-to-3} setting, in which T1, T2, and FLAIR are real and the remaining three modalities are synthesized; (2) a \textbf{2-to-4} setting, in which only T1 and T2 are real; and (3) a \textbf{1-to-5} setting, in which T1 is the sole real input. The three synthesis settings are designed to reflect common modality-missing patterns encountered in routine clinical practice. The 3-to-3 scenario represents typical neuro-oncology protocols where core anatomical sequences are acquired but advanced or contrast-enhanced scans are missing. The 2-to-4 and 1-to-5 settings further simulate more constrained conditions such as emergency imaging or incomplete follow-up examinations.
To eliminate bias, four image sets of each case are presented to the radiologists in a random order, and the readers are blind to the composition of each set. In addition, radiologists review only one image set at a time and take a sufficient time interval between reviewing different sets from the same case, ensuring that the reporting process rely solely on the available imaging information rather than prior knowledge of data provenance.

Table~\ref{tb_reports} presents representative diagnostic reports produced by the two radiologists for selected cases. Each report consists of two components: \emph{Findings}, which systematically describe imaging characteristics such as signal intensity patterns on T1/T2/FLAIR, lesion morphology, location, enhancement features, and diffusion restriction; and \emph{Impression}, which provides diagnostic hypotheses (may include both a tumor grade and specific histologic subtypes) based on the observed radiological evidence. Owing to differences in clinical experience and diagnostic preference, inter-observer variability is expected, and radiologists may generate distinct interpretations for the same case. To quantitatively assess report consistency, we separately evaluate the similarity of the \emph{Findings} and the diagnostic accuracy of the \emph{Impression}. Given the highly structured and templated nature of the \emph{Findings}, direct sentence-level similarity measures using large language models~\cite{bge_embedding} tend to yield inflated similarity scores. Instead, we decompose each \emph{Findings} paragraph into semantically independent clauses (separated by punctuations `,;.'), each corresponding to a specific imaging attribute, and perform bidirectional clause-level matching between \emph{Findings} derived from synthetic sequences and \emph{Findings} derived from real sequences. The final similarity score is computed as the average of the two directional matching results. For \emph{Impression} evaluation, where multiple diagnostic labels may co-exist, we adopt precision, recall, and F1-score to comprehensively assess diagnostic accuracy.

Then, we assess the consistency between radiological reports by analyzing the \emph{Findings} descriptions and the \emph{Impression} diagnoses. Quantitative evaluation of the \emph{Findings} (Table~\ref{tb_meta_analysis_sim}) reveals consistently high textual similarity across all modality-missing scenarios, with similarity scores exceeding 0.8 for both radiologists. This indicates that the synthesized images largely preserve key imaging attributes. Notably, a modest degradation in similarity is observed under the most extreme 1-to-5 setting. In this scenario, deviations mainly manifest as less precise lesion measurements or subtle differences in margin definition and enhancement pattern, which suggests that severe modality missing conditions may introduce localized descriptive variability in detailed image interpretation.
In contrast, analysis of the \emph{Impression} (Table~\ref{tb_meta_analysis_acc_1} and Table~\ref{tb_meta_analysis_acc_2}) demonstrates robustness across different modality-missing conditions. Both radiologists achieve comparable macro- and micro-averaged diagnostic performance regardless of the number of real modalities, with no systematic decline observed in the 1-to-5 setting. This stability indicates that, although fine-grained details in the \emph{Findings} may be affected under severe modality loss, the synthesized images still retain key diagnostic cues relevant to tumor grading and major histological subtypes, supporting consistent radiological impressions.
These results highlight the potential of PMM-Synth to complement incomplete MRI acquisitions and support dependable clinical assessment in real-world imaging workflows.

\subsection*{Ablation Study}
\paragraph{Effectiveness of Personalized Feature Modulation.}
We investigate the effectiveness of PFM through controlled ablation experiments under three training–testing configurations:
(1) \textbf{w/o PFM (single)}, where the backbone Uni-Synth~\cite{zhang2024unified} is trained and evaluated independently on single dataset without PFM, resulting in four separate models corresponding to TTG, TTI, BraTS, and ISLES;
(2) \textbf{w/o PFM (multi)}, where the model is trained on the combined multi-dataset collection without PFM and subsequently evaluated on each individual dataset; and
(3) \textbf{full model}, where the model equipped with PFM is trained on the multi-dataset collection and tested on each dataset individually.
Fig.~\ref{fig_ablation} summarizes the quantitative results. For each dataset, PSNR and SSIM are reported as averages over four synthesis tasks, including two selective one-to-one tasks and two many-to-one tasks.
As shown in Fig.~\ref{fig_ablation}, incorporating PFM consistently improves synthesis performance across all four datasets. In contrast, directly training a unified model on mixed datasets without PFM yields inferior results. This performance degradation can be attributed to inter-dataset heterogeneity, which causes the model to learn averaged feature representations and suppress dataset-specific imaging characteristics.
By explicitly modulating intermediate features conditioned on the dataset source, PFM allows the model to adapt its representations to distinct dataset-specific distributions, thereby improving robustness and generalization to heterogeneous clinical data.

\paragraph{Effectiveness of Modality-Consistent Batch Scheduler.}
Here, we evaluate the effectiveness of MCBS. Our synthesis backbone~\cite{zhang2024unified} adopts multiple modality-specific encoding and decoding streams, which are dynamically activated based on the available modalities of each input. This design requires that all samples within a batch share the same modality availability to ensure consistent stream activation during training.
Without MCBS, randomly sampling batches from multiple datasets often leads to inconsistent modality availability within a batch, restricting the batch size to 1 and severely hindering training efficiency and stability. MCBS overcomes this limitation by constructing batches composed of samples that share the same modality availability and dataset identifier. This strategy enables larger batch sizes without requiring any modifications to the network architecture.
As shown in Fig.~\ref{fig_ablation}, our full model equipped with MCBS ($batch\_size=16$) improves PSNR and SSIM across all four datasets compared with the model without MCBS ($batch\_size=1$). By enforcing intra-batch consistency and allowing for increased batch sizes, MCBS promotes more stable convergence and improves synthesis performance.
In addition to performance gains, MCBS substantially improves training efficiency, reducing the per-epoch training time from 8,724 seconds to 2,078 seconds ($batch_size=16$) on an NVIDIA A100 GPU. Larger batch sizes can be further supported when sufficient GPU memory is available.

\subsection*{Discussion}
In this work, we presents PMM-Synth, a personalized multi-modal MRI synthesis framework that addresses the joint training on multiple clinical datasets. Although existing methods have shown promising results in synthesizing missing modalities within a single dataset, they often struggle to generalize across datasets due to distributional shifts and inconsistent modality coverage commonly encountered in clinical data. PMM-Synth overcomes this limitation by enabling a single model to adapt to diverse datasets and synthesis tasks, thereby achieving both task-level and dataset-level versatility. Our results demonstrate that PMM-Synth consistently surpasses state-of-the-art methods across multiple heterogeneous datasets that vary in disease types, modality coverage, and intensity distributions. Our model achieves superior quantitative and qualitative performance in both one-to-one and many-to-one synthesis scenarios. Besides, incorporating the synthesized modalities into downstream clinical tasks, such as glioma and stroke lesion segmentation, significantly improves the segmentation accuracy. Moreover, blind radiological reporting study with expert radiologists confirm that the synthesized images preserve key radiological features, even in challenging settings where only a single input modality is available. These findings underscore the potential of PMM-Synth to support reliable clinical decision-making when imaging data are incomplete.

The effectiveness of PMM-Synth stems from its capacity to address both inter-dataset distributional shifts and modality inconsistencies. On one hand, the PFM module injects the dataset identifier into the network’s convolutional blocks, enabling the model to learn dataset-specific characteristics and alleviating the adverse effects of inter-dataset distributional shifts during training. On the other hand, the MCBS batching strategy in conjunction with the selective supervision loss addresses modality inconsistencies. MCBS generates larger batches where all samples share the same modality availability to improve training efficiency and convergence. The selective supervision loss further ensures that the model training relies only on available ground truth information.
Despite its strengths, PMM-Synth does not yet address the challenge of continual adaptation to new datasets and unseen modality synthesis tasks after the initial cross-dataset training. In real-world clinical settings, new datasets with novel imaging protocols or additional modalities are frequently introduced. Without careful design, updating the model for such tasks may result in catastrophic forgetting of previously acquired knowledge. Future work could explore two promising directions to mitigate this limitation. First, parameter-efficient tuning techniques, such as Low-Rank Adaptation (LoRA)~\cite{hu2022lora}, could enable the model to learn new tasks using a small number of additional parameters while preserving the core model. Second, integrating continual learning strategies~\cite{wang2024comprehensive} may support lifelong adaptation by allowing the model to incrementally incorporate new information while retaining performance on prior tasks. These extensions would further improve the clinical scalability of PMM-Synth, enabling it to adapt to more complex and diverse clinical datasets.

In conclusion, PMM-Synth provides a versatile solution for multi-modal MRI synthesis across multiple datasets and synthesis tasks. By addressing both inter-dataset distributional shifts and modality inconsistencies, it achieves strong generalization across heterogeneous datasets. Its architectural innovations and superior performance in downstream tasks demonstrate its potential in supporting clinical decision-making. With future extensions toward continual adaptation to more datasets and modalities, PMM-Synth offers a promising direction for scalable deployment in real-world, data-rich clinical environments.

\section*{Methods} 
\subsection*{Ethical issues}
This study was approved by the Institutional Review Board of Beijing Tiantan Hospital, Capital Medical University, Beijing, China (NOs. KY-2019-050-02 and KY-2019-140-02) and ethics committees. Written informed consent was obtained from the participants. The study followed the tenets of the Declaration of Helsinki principles.

\subsection*{Dataset collection and pre-processing}
\paragraph{Dataset.} We use four heterogeneous multi-modal MR datasets, including two private datasets from Beijing Tiantan Hospital (TTG and TTI) and two public datasets (BraTS~\cite{menze2014multimodal,bakas2017advancing,bakas2018identifying} and ISLES~\cite{de2024robust,hernandez2022isles}), to evaluate the proposed method.
As shown in Fig.~\ref{fig_overview}b, these datasets exhibit substantial heterogeneity in modality coverage, disease types, and image intensity distributions.
Specifically, the \textbf{TTG} dataset consists of 983 glioma cases, each comprising six consistently available MR modalities: T1, T2, T1C, FLAIR, DWI, and ADC. Each case includes 24 axial slices.
The \textbf{TTI} dataset contains 246 cases of immune-related diseases. Each case provides a variable subset of T1, T2, FLAIR, and ADC sequences, which exhibits inconsistent modality availability. All cases contain 24 axial slices.
The \textbf{BraTS} dataset includes 292 cases of glioblastoma and lower-grade glioma. Each case offers skull-stripped, co-registered MR images with four consistent modalities: T1, T2, T1C, and FLAIR. Each case consists of 155 axial slices, and is accompanied by binary masks for glioma.
The \textbf{ISLES} dataset comprises 232 ischemic stroke cases. Each includes FLAIR, DWI, and ADC sequences acquired at 1.5T or 3T MRI scanners. All cases are skull-stripped and co-registered. The slice numbers vary between 25 and 73, and voxel-wise binary masks for infarcted areas are provided.

\paragraph{Pre-processing.} All modalities of four datasets are resampled to a consistent voxel spacing, origin, and direction to ensure spatial alignment. For the TTG and TTI datasets, which are not co-registered and skull-stripped, we perform rigid registration to align all modalities to the T1 image for each case, using mutual information as the similarity metric. Besides, brain masks are derived from T1 images using a pre-trained segmentation model and applied across all modalities to remove non-brain regions. Each volume is then center-cropped to a fixed in-plane size of $192\times192$ to focus on anatomical structures.  Finally, we check all processed cases and exclude those with poor registration quality. After pre-processing, 1460, 132, and 161 cases are used for training, validation, and testing, respectively.

\subsection*{Network architecture}
The backbone synthesis network adopts our previously proposed unified synthesis network~\cite{zhang2024unified}.
We extend the network from original four modality to accommodate the modality coverage of multiple datasets.
Given $N$ datasets ($N=4$ in this work), each dataset has its own modality coverage, denoted as $\mathcal{C}_n$ for the $n$-th dataset.  
We compute the union of all available modalities across datasets, i.e., $\mathcal{C} = \bigcup_{n=1}^{N} \mathcal{C}_n$, and expand the synthesis backbone to accommodate all $|\mathcal{C}|$ modalities (T1, T2, T1C, FLAIR, DWI, and ADC in this work).
The updated network consists of one common encoding stream, six modality-specific encoding streams, six decoding streams, and six discriminators, each tailored to process a specific modality. 
Given any subset of the six modalities as input, the model is able to generate the complete set of modalities. 
The framework of our backbone is provided in Supplementary Materials.

Building on this backbone, PMM-Synth further addresses the challenges of distributional shifts and modality inconsistencies in multi-dataset training through three key components: the Personalized Feature Modulation (PFM), Modality-Consistent Batch Scheduler (MCBS), and selective supervision loss. The PFM adaptively adjusts feature processing according to the source dataset, enabling personalized modeling across heterogeneous data and hence avoiding the impact of inter-dataset distributional shifts.
To handle the problem of modality inconsistency, MCBS groups samples with consistent modality availability, allowing for larger batch size and improving training efficiency and stability. Complementarily, the selective supervision loss activates learning only when the target modality is available, ensuring effective learning under incomplete supervision.

\subsection*{Personalized Feature Modulation}
In clinical practice, MR images acquired from different datasets often exhibit distributional shifts due to variations in scanners and acquisition protocols. Naively combining multi-datasets for training can hinder dataset-specific feature modeling, as the network tends to converge toward an averaged representation across datasets. This compromises model's ability to capture distinctive characteristics and leads to suboptimal or even degraded performance on individual datasets.

To address this issue, we introduce a PFM module into the synthesis network. PFM is a plug-and-play component that enables the model to incorporate dataset information and perform differentiated processing for different datasets.
Specifically, for an input image from the $n$-th dataset, we use $n$ as the dataset identifier. The PFM module first applies sinusoidal positional encoding to the identifier, followed by a multi-layer perceptron (MLP) to generate a dataset-specific embedding $D_{\text{emb}}$:
\begin{equation}
D_{\text{emb}} = \text{MLP}(\sin(n)),
\end{equation}
where $\sin(\cdot)$ denotes the sinusoidal encoding function. The resulting embedding $D_{\text{emb}}$ is injected into each convolutional block of the synthesis backbone. Within each block, a new MLP predicts the scale parameter $\gamma$ and shift parameter $\beta$ for feature modulation:
\begin{equation}
\gamma,\ \beta = \text{chunk}(\text{MLP}(D_{\text{emb}})),
\end{equation}
where $\text{chunk}(\cdot)$ splits the output tensor into two equal parts. Feature modulation is then performed via an affine transformation:
\begin{equation}
\hat{F} = F \cdot (\gamma + 1) + \beta,
\end{equation}
where $F$ denotes the input feature the convolutional block, and $\hat{F}$ is the modulated output. By injecting the dataset embedding throughout all convolutional blocks, PFM enables dataset-specific feature adaptation while introducing only a marginal increase in model complexity.

\subsection*{Modality-Consistent Batch Scheduler}
In the multi-dataset training setting, randomly sampled batches from different datasets commonly contain samples with inconsistent modalities. Such inconsistency poses a challenge for batch training. Since our backbone network dynamically selects synthesis streams based on the available modalities of each input, all samples within a mini-batch must share the same modality availability. When this condition is not met, the batch size must be reduced to one, which severely limits training efficiency and hinders convergence.
A naive solution is to adopt per-sample processing using loops or multiple branches accounting for each sample within a batch. However, such implementations are inefficient and significantly increase model complexity.

To overcome these limitations, PMM-Synth introduces the MCBS, a batching strategy that enforces consistent modality availability within each mini-batch. MCBS enables larger batch sizes without modifying the network architecture, thereby improving training efficiency and stability.
To construct modality-consistent batches, MCBS performs the following steps:
\begin{description}
    \item[Step 1: Sample grouping.] 
    For each sample, we record its dataset identifier $n$ and modality availability vector $M = \{m_i\}_{i=1}^6$, where $m_i = 1$ if the $i$-th modality is available and $0$ otherwise. Samples with the same dataset identifier and identical modality availability are then assigned to the same group. 
    \item[Step 2: Group padding.] For each group, if the number of samples is not divisible by the predefined batch size $B$, additional samples are randomly duplicated from the same group to make the total count a multiple of $B$. 
    \item[Step 3: Shuffling.] Within each group, the samples are randomly shuffled and then divided into non-overlapping batches of size $B$. Finally, all batches from all groups are combined and randomly shuffled again to form the final training sequence.
\end{description}

\subsection*{Selective Supervision Loss}
Another challenge arising from inconsistent modalities of multi-datasets is the lack of ground truth for some synthesized modalities during training. To address this, PMM-Synth introduces a selective supervision loss that selectively applies supervision based on the availability of ground truth.

Given a multi-modal MRI sample $Y = \{y_i\}_{i=1}^{6}$, where some modalities may be missing (imputed as zero-filled images), we derive its binary modality availability $M = \{m_i\}_{i=1}^{6}$. We further introduce a random synthesis condition $SC = \{sc_i\}_{i=1}^{6}$ to mask out a subset of modalities for training, where $sc_i = 1$ denotes a source modality, and $sc_i = 0$ denotes a target modality (to be synthesized), with the constraint that $sc_i = 0$ when $m_i = 0$. Then, the network input $X = \{x_i\}_{i=1}^{6}$ is constructed as:

\begin{equation}
    x_i = y_i \cdot sc_i
\end{equation}
After feeding $X$ into the generator of PMM-Synth, the outputs $\hat{Y}=\{\hat{y}_i\}_{i=1}^6$ ($\hat{y}_i$ with $sc_i=1$ corresponds to the reconstruction image and $\hat{y}_i$ with $sc_i=0$ corresponds to the synthetic image for imputation) are evaluated by modality-specific discriminators $Dis = \{Dis_i\}_{i=1}^{6}$ to assess realism.

Following our previous work~\cite{zhang2024unified}, the generator loss $L_G$ consists of three terms: synthesis loss $L_{syn}$, reconstruction loss $L_{rec}$, and adversarial loss $L_{adv}$. The key difference lies in the incorporation of a selective supervision mechanism based on $M$ and $SC$:

\begin{equation}
    L_G = \lambda_1 L_{syn} + \lambda_2 L_{rec} + \lambda_3 L_{adv}
\end{equation}


\begin{equation}
    L_{syn} = \sum_{i=1}^{6} (1 - sc_i) \cdot m_i \cdot \left\| \hat{y}_i - y_i \right\|_1
\end{equation}

\begin{equation}
    L_{rec} = \sum_{i=1}^{6} sc_i \cdot m_i \cdot \left\| \hat{y}_i - y_i \right\|_1
\end{equation}


\begin{equation}
    L_{adv} = \sum_{i=1}^{6} (1 - sc_i) \cdot m_i \cdot \left\| Dis_i(\hat{y}_i) - 1 \right\|_2
\end{equation}

Here, $\|\cdot\|_1$ and $\|\cdot\|_2$ denote the L1 and L2 norms. 
$L_{syn}$ measures the difference between synthetic images and real images. The coefficient $(1-sc_i)\cdot m_i$ ensures that only the target modality which has ground truth is selected for computation. $L_{rec}$ measures the difference between reconstruction images and real images. The coefficient $sc_i\cdot m_i$ ensures that only the source modality which has ground truth is selected for computation.
The trade-off coefficients $\lambda_1$, $\lambda_2$, and $\lambda_3$ are set to 100, 30, and 1, respectively.

Likewise, the discriminator loss $L_{D}$ also employs the selective supervision mechanism, where the supervision is activated only when the corresponding ground truth for the synthesized target modality is available:

\begin{equation}
    L_{D} = \sum_{i=1}^{6} (1 - sc_i) \cdot m_i \cdot \left( \left\| Dis_i(\hat{y}_i) \right\|_2 + \left\| Dis_i(y_i) - 1 \right\|_2 \right)
\end{equation}

This selective supervision mechanism ensures that only modalities with available ground truth contribute to the loss, which enables flexible and effective learning under incomplete modality conditions.

\subsection*{Evaluation metrics}
In this study, we use Peak Signal-to-Noise Ratio (PSNR) and Structural Similarity Index Measure (SSIM) as quantitative metrics to evaluate the quality of the synthesized images.

\paragraph{Peak Signal-to-Noise Ratio (PSNR).} PSNR is a commonly used metric to evaluate the pixel-wise similarity between a synthesized image $\hat{y}$ and its corresponding ground truth $y$, with higher PSNR values indicating better image fidelity. PSNR is defined as:
\begin{equation}
    \mathrm{PSNR} = 10 \cdot \log_{10}\left( \frac{MAX^2}{\mathrm{MSE}} \right),
\end{equation}
where $MAX$ is the maximum pixel value of the image, and $\mathrm{MSE}$ denotes the mean squared error between $\hat{y}$ and $y$.

\paragraph{Structural Similarity Index (SSIM).} SSIM evaluates the perceptual similarity between two images by considering luminance, contrast, and structural information. A higher SSIM values reflect better perceived visual quality. SSIM is defined as:
\begin{equation}
    \mathrm{SSIM}(y, \hat{y}) = \frac{(2\mu_y \mu_{\hat{y}} + C_1)(2\sigma_{y\hat{y}} + C_2)}{(\mu_y^2 + \mu_{\hat{y}}^2 + C_1)(\sigma_y^2 + \sigma_{\hat{y}}^2 + C_2)},
\end{equation}
where $\mu_y$ and $\mu_{\hat{y}}$ are the mean intensities of the ground truth and synthesized images, $\sigma_y^2$ and $\sigma_{\hat{y}}^2$ are their variances, and $\sigma_{y\hat{y}}$ is the covariance. $C_1$ and $C_2$ are small constants to stabilize the division.

\section*{Data availability}
Two private brain MRI datasets TTG and TTI originate from clinical routine, protected by national privacy laws and hence are not publicly available or shareable with third parties due to the retrospective nature of the study performed and the associated IRB approval for the study limiting the data use.
The public BraTS dataset can be accessed from \url{https://www.med.upenn.edu/cbica/brats-2019/}. The ISLES dataset can be accessed from \url{https://isles22.grand-challenge.org/}.


\small
\bibliographystyle{unsrt}
\bibliography{reference}

\normalsize

\section*{Acknowledgements}
This research was supported by Natural Science Foundation of China under Grant 62271465, Suzhou Basic Research Program under Grant SYG202338, the Key Program of the National Natural Science Foundation of China under Grant 82330057, and Beijing Hospital Management Center-Climb Plan under Grant DFL20220503.

\section*{Author contributions}
In this work, Y. Z., Y. L., and S. K. Z. conceived the study and designed the experiments. Y. Z. performed the experiments, conducted the data analysis, and drafted the initial manuscript. Z. Z., S. X., S. L., Z. L., and J. Q. contributed to data curation and data management. Z. Z., and S. X. designed the radiological reporting experiments. S. L., Z. L., and J. Q. were responsible for radiological assessment and preparation of radiological reports. Y. Z., Z. Z., Q., W., Y. L., and S. K. Z. critically revised the manuscript.

\section*{Competing interests}
The authors declare no competing interests.

\clearpage
\setcounter{figure}{0}
\renewcommand{\thefigure}{S\arabic{figure}}

\section*{Supplementary information}
\begin{figure*}[h]
    \centering
    \includegraphics[width=\textwidth]{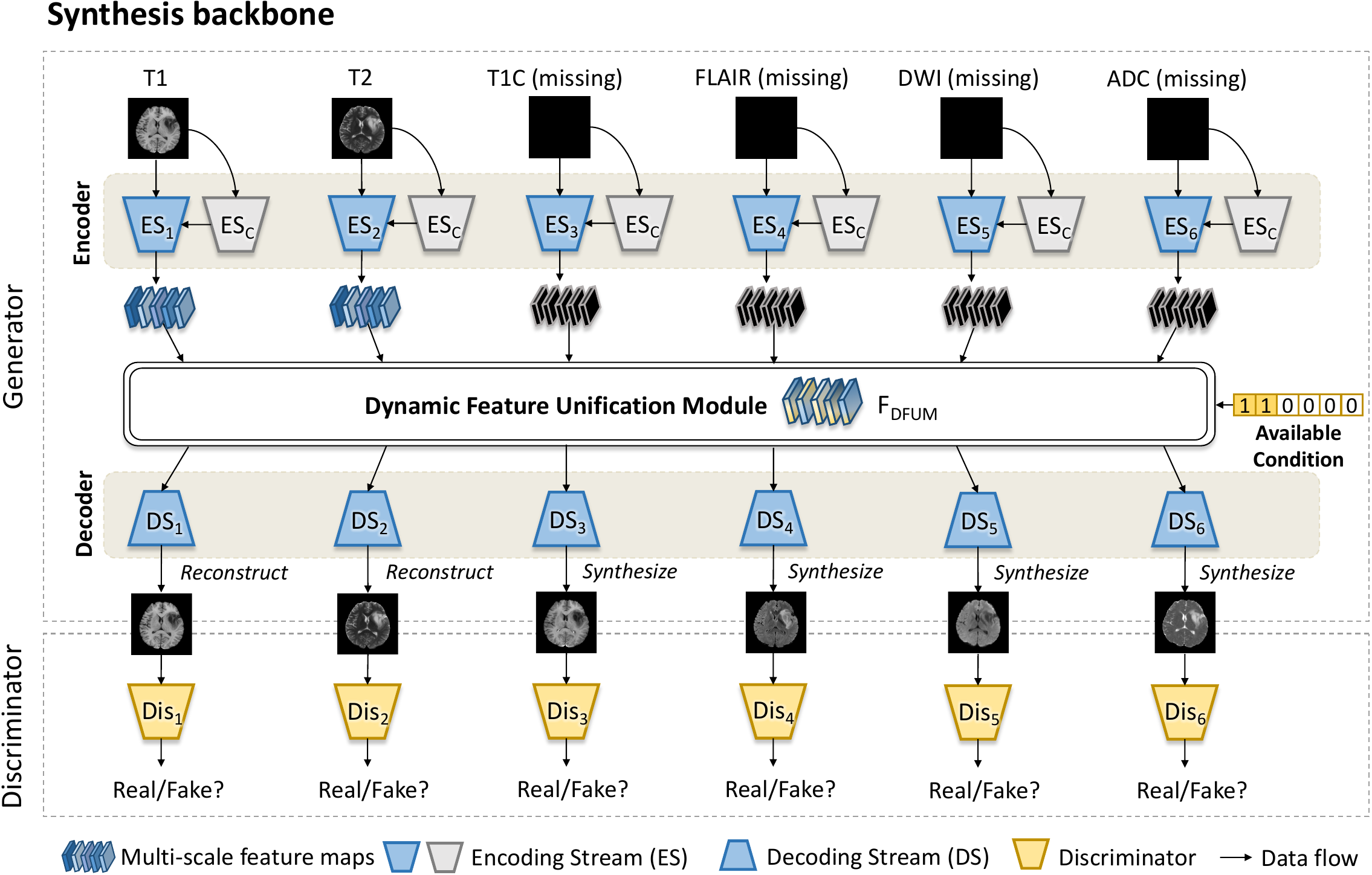}
\caption{
The synthesis backbone adopted in this work is extended from our previous study. It comprises one common encoding stream, six modality-specific encoding streams, six decoding streams, and six discriminators, each specifically designed to process a corresponding imaging modality. A DFUM module is employed to integrate feature representations from a variable number of input modalities.} 
\end{figure*}

\end{document}